\documentstyle[twoside,cl]{article}

\issue{00}{0}{1998}

\title{Practical Experiments with Regular Approximation of Context-free
Languages}

\author{ Mark-Jan Nederhof%
 \thanks{DFKI, Stuhlsatzenhausweg 3, D-66123 Saarbr{\"u}cken, Germany.
E-mail: nederhof@dfki.de} \\
\affil{German Research Center for Artificial Intelligence}
}

\runningtitle{Experiments with Regular Approximation}
\runningauthor{Nederhof}

\input{psfig}

\newcommand{\myterm}{{\it\Sigma}}

\newcommand{\myvar}{V}

\newtheorem{Definition}{Definition}

\newcommand{\bul}{\mathrel{\bullet}}

\newcommand{\de}{\rightarrow}
\newcommand{\dm}{\de^{*}}

\newcommand{\ep}{\epsilon}

\newcommand{\cF}{{\cal F}}

\newcommand{\procedure}{\mbox{\bf procedure\ }}
\newcommand{\return}{\mbox{\bf return\ }}
\newcommand{\DO}{\mbox{\bf do\ }}
\newcommand{\LET}{\mbox{\bf let\ }}
\newcommand{\IF}{\mbox{\bf if }}
\newcommand{\THEN}{\mbox{\bf then }}
\newcommand{\ELSE}{\mbox{\bf else }}
\newcommand{\ELSEIF}{\mbox{\bf elseif }}
\newcommand{\END}{\mbox{\bf end}}
\newcommand{\foreach}{\mbox{\bf for each }}

\newcommand{\suchthat}{\mbox{\bf such that }}
\newcommand{\some}{\mbox{\bf some }}

\newcommand{\freshstate}{\mbox{\it fresh\_state\/}}
\newcommand{\makefsa}{\mbox{\it make\_fa\/}}

\newcommand{\recursive}{\mbox{\it recursive\/}}

\newcommand{\Left}{\mbox{\it LeftGenerating\/}}
\newcommand{\Right}{\mbox{\it RightGenerating\/}}

\newcommand{\LEFT}{\mbox{\it left\/}}
\newcommand{\RIGHT}{\mbox{\it right\/}}
\newcommand{\cyclic}{\mbox{\it cyclic\/}}
\newcommand{\self}{\mbox{\it self\/}}

\newcommand{\spaceA}{\ \ }

\newtheorem{ex1}{Example}

\begin{document}

\maketitle

\begin{abstract}
Several methods are discussed that construct a finite automaton given 
a context-free grammar, including both methods that lead to subsets and
those that lead to supersets of the original context-free language.
Some of these methods of {\bf regular approximation\/} are new,
and some others are presented here in a more refined form with 
respect to existing literature.
Practical experiments with the different methods of regular approximation are performed
for spoken-language input:
hypotheses from a speech recognizer
are filtered through a finite automaton.
\end{abstract}

\section{Introduction}

Several methods of regular approximation of context-free languages
have been proposed in the literature. 
For some, the regular language is a superset of the context-free language, and
for others it is a subset.
We have implemented a large number of methods, 
and where needed we
refined them with an analysis of the grammar. We also propose a number of new
methods.

The analysis is based on a
sufficient condition for context-free grammars to generate regular languages.
For an arbitrary grammar,
this analysis identifies sets of rules that need to be processed in a special way in 
order to obtain a regular language. The nature of this processing differs for the 
respective approximation methods. For other parts of the grammar, no special
treatment is needed and the grammar rules are translated to states and transitions
of a finite automaton without affecting the language.

Few of the published articles on regular approximation have discussed
the application in practice. In particular, little
attention has been given to the following two questions. 
First, what happens when a context-free grammar grows in size? What is then the
increase of the sizes of the
intermediate results and the obtained minimal deterministic automaton?
Second, how ``precise'' are the approximations? That is, how much larger than
the original context-free language is
the language obtained by a superset approximation, and how much smaller is
the language obtained by a subset approximation? (How we measure the ``sizes''
of languages in a practical setting will become clear in the sequel.)

Some considerations with regard to theoretical upper bounds on the sizes of the intermediate
results and the finite automata have already been discussed by \namecite{NE97}.
In this article we will try to answer the above two questions 
in a practical setting, using 
practical linguistic grammars and sentences taken from a spoken-language corpus.

The structure of this paper is as follows. In Section~\ref{prel} we
recall some standard definitions from language theory. 
Section~\ref{trees} investigates a sufficient condition for a
context-free grammar to
generate a regular language. We also present the construction of a
finite automaton from such a grammar.

In Section~\ref{approx}, we discuss
several methods to approximate the language generated by a grammar if the
sufficient condition mentioned above is not satisfied. These methods
can be enhanced by a grammar transformation presented in Section~\ref{increase}.
Section~\ref{compare} compares
the respective methods, which leads to
conclusions in Section~\ref{conclude}.

\section{Preliminaries}
\label{prel}

Throughout this paper we use standard formal language notation
(see e.g.\ \namecite{HA78}). In this section we recall some 
basic definitions.

A {\bf context-free grammar\/} $G$ is a $4$-tuple
$(\myterm,N,P,S)$, where $\myterm$ and $N$
are two finite disjoint sets of terminals
and nonterminals, respectively, $S \in N$ is the start symbol,
and $P$ is a finite set of rules. Each rule has the form
$A \de \alpha$ with $A \in N$ and $\alpha \in \myvar^*$,
where $\myvar$ denotes $N \cup \myterm$. The relation $\de$ 
on $N\times \myvar^*$ is
extended to a relation on $\myvar^*\times\myvar^*$ as usual.
The transitive and reflexive closure of $\de$ is denoted by $\dm$.

The language {\bf generated\/} by $G$ is given by
the set
$\{w\in\myterm^*\ |\ S\dm w\}$. By definition, such a set is a 
{\bf context-free\/} language.
By {\bf reduction\/} of a grammar we mean the elimination from
$P$ of all rules $A\de \gamma$ 
such that $S\dm \alpha A \beta \de \alpha \gamma \beta \dm w$ does not hold for any
$\alpha,\beta\in \myvar^*$ and $w\in\myterm^*$.

We generally use symbols $A, B, C, \ldots$ to range over $N\!$,
symbols $a,b,c, \ldots$ to range over $\myterm\!$,
symbols $X,Y,Z$ to range over $\myvar\!$,
symbols $\alpha, \beta, \gamma, \ldots$ to range over $\myvar^*\!$,
and symbols $v, w, x, \ldots$ to range over $\myterm^*\!$.
We write $\ep$ to denote the empty string.


A rule of the form $A\de B$ is called a {\bf unit\/} rule.

A (nondeterministic) {\bf finite automaton\/} $\cF$ is a 5-tuple
$(K,\myterm,\Delta,s,F)$, where $K$ is a finite set of {\bf states\/},
of which $s$ is the {\bf initial\/} state and those in $F\subseteq K$ 
are the {\bf final\/} states,
$\myterm$ is the input alphabet,
and the {\bf transition relation\/} $\Delta$ 
is a finite subset of $K\times \myterm^* \times K$. 

We define
a {\bf configuration\/} to be an element of $K\times \myterm^*$.
We define the binary relation $\vdash$ between configurations as:
$(q,vw)\vdash(q',w)$ if and only if $(q,v,q')\in\Delta$.
The transitive and reflexive closure of $\vdash$ is denoted by $\vdash^*$.

Some input $v$ is {\bf recognized\/} if $(s,v)\vdash^*(q,\ep)$, for some
$q\in F$.
The language
{\bf accepted\/} by $\cF$ is defined
to be the set of all strings $v$ that are recognized.
By definition, a language accepted by a finite automaton is called
a {\bf regular\/} language.

\section{Finite Automata in Absence of Self-embedding}
\label{trees}



We define a {\bf spine\/} in a parse tree to be a path that runs from
the root down to some leaf. 
Our main interest in spines lies in the sequences of 
grammar symbols at nodes bordering on spines. 

\begin{figure*}[tb]
\hspace*{\fill}
\psfig{figure=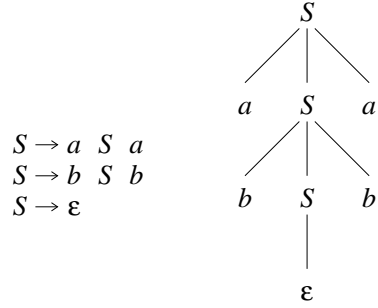}
\hspace*{\fill}
\caption{Grammar of palindromes, and a parse tree.}
\label{pal}
\end{figure*}

A simple example is the set of
parse trees such as the one in Figure~\ref{pal}, for a grammar
of palindromes. 
It is intuitively
clear that the language is not regular: the grammar symbols to the
left of the spine from the root to $\ep$ ``communicate'' 
with those to the right of the spine.
More precisely, the prefix of the input up to the point where it meets
the final node $\ep$ of the spine
determines the suffix after that point, in
a way that an unbounded quantity of symbols from the prefix need to be
taken into account. 

A formal explanation for why the grammar may not generate a
regular language relies on the following definition
\cite{CH59}:
\begin{Definition}
A grammar is {\bf self-embedding\/} if 
there is some $A\in N$ such that $A\dm \alpha A \beta$, for
some $\alpha\neq\ep$ and $\beta\neq\ep$.
\end{Definition}
If a grammar is not self-embedding, this
means that when a section
of a spine in a parse tree repeats itself, then either no grammar symbols 
occur to the left of that section of the spine, 
or no grammar symbols occur to the right.
This prevents the ``unbounded communication'' between the two sides 
of the spine exemplified by the palindrome grammar.

We now prove that grammars that are not self-embedding
generate regular languages.
For an arbitrary grammar, 
we define the set of {\bf recursive\/} nonterminals as:
$$
\overline{N} = \{A\in N\ |\ \exists \alpha,\beta[A \dm \alpha A\beta]\} 
$$
We determine the partition ${\cal N}$ of $\overline{N}$ consisting of
subsets $N_1,N_2,\ldots,N_k$, for some $k\geq 0$, of {\bf mutually 
recursive\/} nonterminals:
\begin{center}
${\cal N} = \{N_1,N_2,\ldots,N_k\}$ \\[.1pt]
$N_1\cup N_2 \cup \cdots \cup N_k = \overline{N} $\\[.1pt]
$\forall i[N_i \neq \emptyset]\mbox{ and } 
\forall i,j[i\neq j \Rightarrow N_i \cap N_j=\emptyset]$ \\[.1pt]
$\exists i[A \in N_i \wedge B \in N_i]\ \  \Leftrightarrow \ \ 
\exists \alpha_1,\beta_1,\alpha_2,\beta_2[A \dm \alpha_1 B\beta_1 
\wedge B \dm \alpha_2 A \beta_2], \mbox{ for all } A,B\in \overline{N} $
\end{center}
We now define the function $\recursive$ from ${\cal N}$ to the set
$\{\LEFT,\RIGHT,\self,\cyclic\}$. For $0\leq i \leq k$:
$$
\begin{array}[t]{rclcrcr}
\recursive(N_i) &=& \LEFT, & {\it if }&\neg\Left(N_i)&\wedge&\Right(N_i) \\
                &=& \RIGHT, & {\it if }&\Left(N_i)&\wedge&\neg\Right(N_i) \\
                &=& \self, & {\it if }&\Left(N_i)&\wedge&\Right(N_i) \\
                &=& \cyclic, & {\it if }&\neg\Left(N_i)&\wedge&\neg\Right(N_i)
\end{array}
$$
where
\begin{eqnarray*}
\Left(N_i) &=& \exists (A \de \alpha B \beta)\in P
[A\in N_i \wedge B\in N_i \wedge \alpha\neq\ep] \\
\Right(N_i) &=& \exists (A \de \alpha B \beta)\in P
[A\in N_i \wedge B\in N_i \wedge \beta\neq\ep]
\end{eqnarray*}
When $recursive(N_i) = \LEFT$, $N_i$ consists of only left-recursive 
nonterminals,
which does not mean it cannot also contain right-recursive nonterminals,
but in that case right recursion amounts to application of unit rules.
When $recursive(N_i) = \cyclic$, 
it is {\em only\/} such unit rules that take part in the recursion.

That $recursive(N_i) = \self$, for some $i$, is a sufficient and necessary
condition for the grammar to be 
self-embedding. 
Therefore, we have to prove that if
$recursive(N_i) \in \{\LEFT,\RIGHT,\cyclic\}$, for all $i$, then the grammar
generates a regular language.
Our proof 
differs from an existing proof \cite{CH59a} in that
it is fully constructive: Figure~\ref{toFS} presents
an algorithm for creating a finite automaton that accepts the
language generated by the grammar.

\begin{figure*}[p]
\begin{center}
\begin{tabbing}
\LET $K$ = $\emptyset$,\
$s$ = \freshstate,\
$f$ = \freshstate,\
$F$ = $\{f\}$; \\
\makefsa$(s,S,f)$.  \\
\\
\procedure \makefsa$(q_0,\alpha,q_1)$: \\
\spaceA\= \IF $\alpha=\epsilon$  \\
\> \THEN \LET $\Delta$ = $\Delta\cup\{(q_0,\epsilon,q_1)\}$ \\
\> \ELSEIF $\alpha=a$, \some $a\in\myterm$  \\
\> \THEN \LET $\Delta$ = $\Delta\cup\{(q_0,a,q_1)\}$ \\
\> \ELSEIF $\alpha=X\beta$, \some $X\in V$,
        $\beta\in V^*$ \suchthat $|\beta| > 0$ \\
\> \THEN\= \LET $q$ = \freshstate; \\
\> \> \makefsa$(q_0,X,q)$; \\
\> \> \makefsa$(q,\beta,q_1)$ \\
\> \ELSE\= \LET $A$ = $\alpha$;\ \ \
        (* $\alpha$ must consist of a single nonterminal *) \\
\>\> \IF {\bf there exists} $i$ {\bf such that} $A\in N_i$ \\
\>\> \THEN\=  \foreach $B\in N_i$ \DO \LET $q_B$ = \freshstate\ \END; \\
\> \>\>  \IF \recursive$(N_i)$ = \LEFT \\
\> \>\>  \THEN\= \foreach $(C\de X_1\cdots X_m)\in P$ \suchthat
                   $C\!\in\! N_i\wedge X_1,\ldots, X_m \!\notin\! N_i$  \\
\> \>\>  \> \DO \makefsa$(q_0,X_1\cdots X_m, q_C)$ \\
\> \>\>  \> \END; \\
\> \>\>  \> \foreach\= $(C\de DX_1\cdots X_m)\in P$ \suchthat \\
\> \>\>  \> \> \ \ \
                   $C,D\in N_i\wedge X_1,\ldots, X_m \notin N_i$  \\
\> \>\>  \> \DO \makefsa$(q_D, X_1\cdots X_m, q_C)$ \\
\> \>\>  \> \END; \\
\> \>\>  \> \LET $\Delta$ = $\Delta\cup\{(q_A,\epsilon,q_1)\}$ \\
\> \>\>  \ELSE\= (* \recursive$(N_i)$ $\in$ $\{\RIGHT,\cyclic\}$ *) \\
\> \>\>  \> \foreach $(C\de X_1\cdots X_m)\in P$ \suchthat
                   $C\!\in\! N_i\wedge X_1,\ldots, X_m\! \notin\! N_i$  \\
\> \>\>  \> \DO \makefsa$(q_C,X_1\cdots X_m, q_1)$ \\
\> \>\>  \> \END; \\
\> \>\>  \> \foreach\= $(C\de X_1\cdots X_mD)\in P$ \suchthat \\
\> \>\>  \> \> \ \ \ 
                   $C,D\in N_i\wedge X_1,\ldots, X_m \notin N_i$  \\
\> \>\>  \> \DO \makefsa$(q_C, X_1\cdots X_m, q_D)$ \\
\> \>\>  \> \END; \\
\> \>\>  \> \LET $\Delta$ = $\Delta\cup\{(q_0,\epsilon,q_A)\}$ \\
\> \>\>  \END \\
\>\>  \ELSE \foreach\= $(A\de \beta)\in P$ \DO \makefsa$(q_0, \beta, q_1)$ \END\ \ \
               (* $A$ is not recursive *) \\
\>\>  \END \\
\>  \END \\
\END. \\
\\
\procedure \freshstate$()$: \\
\> create some object $q$ such that $q\notin K$; \\
\> \LET $K$ = $K\cup\{q\}$; \\
\> \return $q$ \\
\END.
\end{tabbing}
\end{center}
\caption{Transformation from a grammar $G=(\myterm,N,P,S)$ that is not
self-embedding
into an equivalent finite automaton $\cF=(K,\myterm,\Delta,s,F)$.}
\label{toFS}
\end{figure*}

The process is initiated at the start symbol, and from there the
process descends the grammar
in all ways until terminals are encountered, and then transitions
are created labelled with those terminals.
Descending the grammar is
straightforward in the case of rules of which the left-hand side is not a
recursive nonterminal: the subautomata found recursively for 
members in the right-hand side will be connected.
In the case of recursive nonterminals, the process depends
on whether the nonterminals in the corresponding set from ${\cal N}$
are mutually left-recursive or right-recursive; if they are both, which
means they are cyclic, then
either subprocess can be applied; in the code in Figure~\ref{toFS}
cyclic and right-recursive subsets $N_i$ are treated uniformly.

We discuss the case that the nonterminals are left-recursive.
One new state is created for each nonterminal in the set. The transitions
that are created for terminals and nonterminals not in $N_i$ are
connected in a way that is reminiscent of the construction of
left-corner parsers \cite{RO70},
and specifically of one construction
that focuses on sets of mutually recursive nonterminals 
\cite[Section~5.8]{NE94b}.

\begin{figure*}[tb]
\hspace*{\fill}
\begin{minipage}[b]{2.6cm}
\begin{eqnarray*}
S &\de& {\it Aa} \\
A &\de& {\it SB} \\
A &\de& {\it Bb} \\
B &\de& {\it Bc} \\
B &\de& d
\end{eqnarray*}
\end{minipage}
\ \ \ \ \ 
\begin{minipage}[b]{7.4cm}
$\overline{N} = \{S,A,B\}$ \\
${\cal N} = \{N_1,N_2\}$ \\
\makebox[2cm][l]{$N_1=\{S,A\}$} $\recursive(N_1) = \LEFT$ \\
\makebox[2cm][l]{$N_2=\{B\}$} $\recursive(N_2) = \LEFT$  \\
\end{minipage}
\hspace{\fill} \\
\psfig{figure=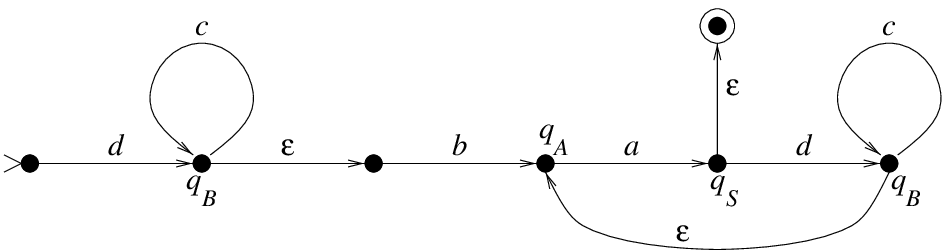}
\hspace*{\fill}
\caption{Application of the code from Figure~\protect\ref{toFS}
on a small grammar.}
\label{fsa}
\end{figure*}

An example is given in Figure~\ref{fsa}. Four states have been labelled
according to the names they are given in procedure $\makefsa$. There
are two states that are labelled $q_B$. This can be explained by the fact that
nonterminal $B$ can be reached by descending the grammar from $S$
in two essentially distinct ways.

The code in Figure~\ref{toFS} differs from the 
actual implementation in that sometimes 
for a nonterminal a separate finite automaton is constructed,
viz.\ 
for those nonterminals that occur as $A$ in the code.
A transition in such a subautomaton may be labelled by another nonterminal 
$B$, which then
represents the subautomaton corresponding to $B$.
The resulting representation is similar to extended context-free grammars
\cite{PU81}, with the exception that in our case recursion cannot occur,
by virtue of the construction.

\begin{figure*}[tb]
\begin{center}
\psfig{figure=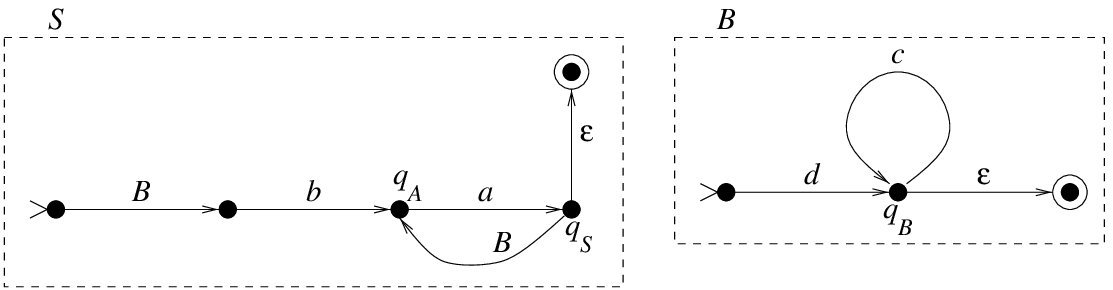}
\end{center}
\caption{The automaton from Figure~\protect\ref{fsa} in a
compact representation.}
\label{mac}
\end{figure*}

The representation for the running example is indicated by Figure~\ref{mac}, which
shows two subautomata, labelled $S$ and $B$. The one labelled $S$ is the
automaton on the top level, and contains two transitions labelled $B$,
which refer to the other subautomaton. 
Note that this representation is more compact than the one from 
Figure~\ref{fsa}, since the transitions that are involved in representing the 
sublanguage of strings generated by nonterminal $B$ are included only once.

The compact representation consisting of subautomata
can be turned into a single finite automaton by substituting subautomata
$A$ for transitions labelled $A$ in other automata. This comes down
to regular substitution in the sense of \namecite{BE79}.
The advantage of this way of obtaining a finite automaton 
over a direct construction of
a nondeterministic automaton is that subautomata may be determinized
and minimized before they are substituted into larger subautomata.
Since in many cases determinized and minimized automata are much
smaller, this process avoids much of the combinatorial explosion
that takes place upon naive construction of a single nondeterministic 
finite automaton.%
\footnote{%
The representation by \namecite{MO98} is even more compact
than ours for grammars that are not self-embedding. However,
in the sequel we are going to use our representation 
as intermediate result in
approximating an unrestricted context-free grammar, with the 
final objective of obtaining a single minimal deterministic automaton. For this
purpose, the representation by \namecite{MO98}
offers little advantage.}

Assume we have a list of subautomata $A_1,\ldots,A_m$
that is ordered from lower level to higher level automata; i.e.\ 
if an automaton $A_p$ occurs as label of a transition of 
automaton $A_q$, then $p<q$; $A_m$ must be the start symbol $S$. 
This order is a natural result of the way that subautomata are constructed
during our depth-first traversal of the grammar, 
which is actually {\bf postorder} in the sense that a subautomaton is output
after all subautomata occurring at its transitions have been
output.

Our implementation constructs a minimal
deterministic automaton by repeating the following for
$p=1,\ldots,m$:
\begin{enumerate}
\item\ Make a copy of $A_p$. Determinize and minimize the copy. If it has
fewer transitions labelled by nonterminals than the original, then
replace $A_p$ by its copy.
\item\ Replace each transition in  $A_p$ of the form $(q,A_r,q')$ by
(a copy of) automaton $A_r$ in a straightforward way. 
This means that new $\ep$-transitions
connect $q$ to the start state of $A_r$ and the
final states of $A_r$ to $q'$.
\item\ Again determinize and minimize $A_p$ and store it for later reference.
\end{enumerate}
The automaton obtained for $A_m$ after step~3 is the desired result.


\section{Methods of Regular Approximation}
\label{approx}

This section describes a number of methods for approximating
a context-free grammar by means of a finite automaton. Some published 
methods did not mention self-embedding explicitly 
as potential source of non-regularity of the language, and
suggested that approximations should be applied
globally for the complete grammar.
Where this is the case, we adapt the method so that
it is more selective and deals with self-embedding locally.

The approximations
are integrated into the construction of the finite automaton from the grammar,
which was described in the previous section.
A separate
incarnation of the approximation process is activated upon finding a nonterminal $A$ such that
$A\in N_i$ and 
$\recursive(N_i)=\self$, for some $i$. This incarnation then only pertains to 
the set of rules of the form $B\de\alpha$, where $B\in N_i$. 
In other words, nonterminals not in $N_i$ are treated by this
incarnation of the approximation process as if they were terminals.

\subsection{Superset approximation based on RTNs}
\label{approx:ned}

The following approximation was proposed by \namecite{NE97}. The presentation here
however differs substantially from the earlier publication,
which treated the approximation process entirely on the level of context-free grammars:
a self-embedding grammar was transformed in such a way that it was no longer 
self-embedding. A finite automata was then obtained from the
grammar by the algorithm discussed above.

The presentation here is based on recursive transition networks
(RTNs) \cite{WO70a}. We can see a context-free
grammar as an RTN as follows.
We introduce two states $q_A$ and $q_A'$
for each nonterminal $A$, and $m+1$ states $q_0,\ldots, q_m$
for each rule $A\de X_1\cdots X_m$. The states for a rule $A\de X_1\cdots X_m$
are connected with each other and to the states for the left-hand side $A$ by 
one transition $(q_A,\ep,q_0)$, a transition
$(q_{i-1},X_i,q_i)$ for each $i$ such that $1\leq i\leq m$,
and one transition $(q_m,\ep,q_A')$. (Actually, some epsilon transitions are avoided in
our implementation, but we will not be concerned with such optimizations here.)

In this way, we obtain a finite automaton with initial state $q_A$ and final state $q_A'$
for each nonterminal $A$ and its defining rules 
$A\de X_1\cdots X_m$.
This automaton can be seen as one component of the RTN. The complete RTN
is obtained by the collection of all such finite automata for 
different nonterminals.

\begin{figure*}[tb]
\begin{center}
\psfig{figure=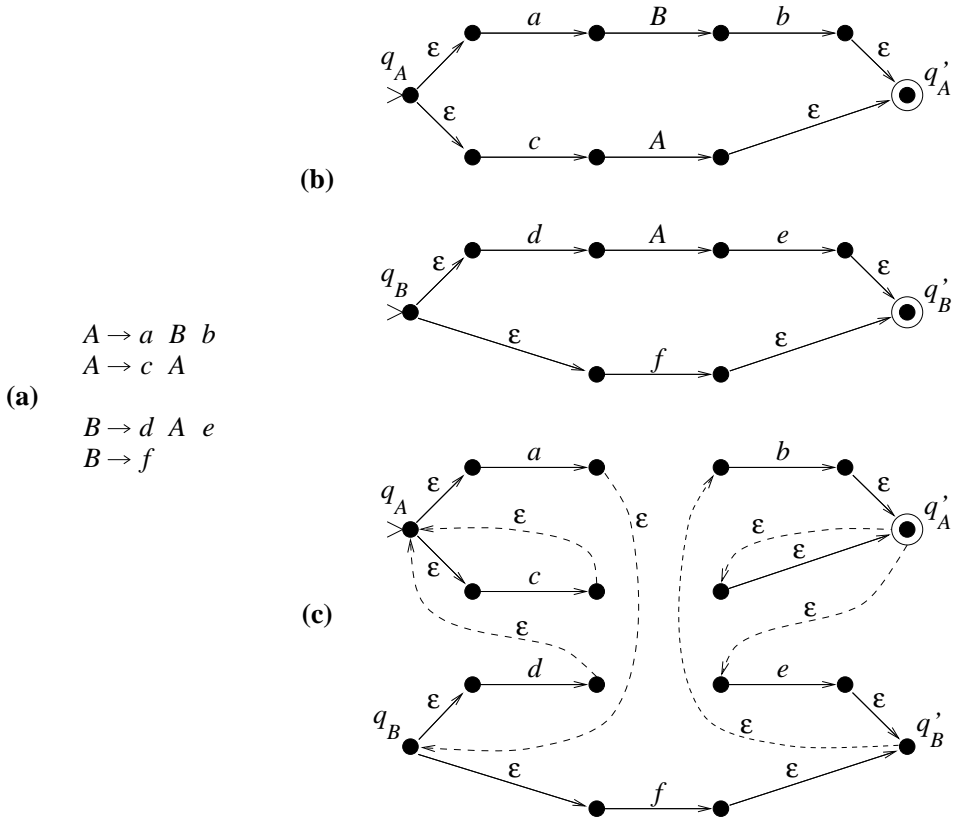}
\end{center}
\caption{Application of the RTN method for the grammar at (a). The RTN is
given at (b), and (c) presents the approximating finite automaton. We assume
$A$ is the start symbol and therefore $q_A$ becomes initial state and
$q_A'$ becomes final state in the approximating automaton.}
\label{rtn}
\end{figure*}

An approximation now results if we join all the components in
one big automaton, and if we approximate the usual mechanism of recursion
by replacing each transition $(q,A,q')$ by two transitions
$(q,\ep,q_A)$ and $(q_A',\ep,q')$. The construction is 
illustrated in Figure~\ref{rtn}.

In terms of the original
grammar, this approximation can be informally explained as follows.
Suppose we have three rules $B \de \alpha A \beta$,
$B' \de \alpha' A \beta'$, and $A\de\gamma$. Top-down left-to-right
parsing would proceed for example by recognizing $\alpha$ in the first
rule; it would then descend into rule $A\de\gamma$, and recognize $\gamma$;
it would then return to the first rule and subsequently process $\beta$.
In the approximation however, the finite automaton ``forgets'' which rule it came from
when it starts to recognize $\gamma$, so that it may subsequently 
recognize $\beta'$ in the second rule.

For the sake of presentational convenience, the above describes 
a construction working on the complete grammar. However, 
our implementation applies the construction separately for each
nonterminal in a 
set $N_i$ such that $\recursive(N_i)=\self$, which leads to a separate
subautomaton of the compact representation (Section~\ref{trees}).

See \namecite{NE98c} for a variant of this approximation that
constructs finite transducers rather than finite automata.

We have further implemented a parameterized version of the RTN
approximation. A state of the nondeterministic automaton
is now also associated to a list $H$ of length $|H|$ strictly
smaller than a number
$d$, which is the parameter to the method. This list
represents a history of
rule positions that were encountered in the computation leading to the
present state.

More precisely, we define an {\bf item\/} to be an object of the form
$[A\de\alpha\bul\beta]$, where $A\de\alpha\beta$ is a rule from the grammar.
These are the same objects as the ``dotted'' productions from 
\namecite{EA70}. The dot indicates a position in the right-hand side.

The unparameterized RTN method had one state 
$q_I$ for each item $I$, and
two states $q_A$ and $q'_A$ for each nonterminal $A$. The parameterized
RTN method has one state $q_{\it IH}$ for each item $I$ and
each list of items $H$ that represents a valid history for reaching $I$, and two
states $q_{\it AH}$ and $q'_{\it AH}$ for each nonterminal $A$ and 
each list of items $H$ that represents a valid history for reaching $A$. 
Such a valid history is defined to be a list $H$ with $0\leq |H|<d$ that represents
a series of positions in rules that could have been invoked before reaching $I$ or $A$,
respectively.
More precisely,
if we set $H=I_1\cdots I_{n}$, then each $I_m$ 
($1\leq m \leq n$) should be of the form 
$[A_m\de \alpha_m \bul B_m\beta_m]$ and for $1\leq m < n$ we should have
$A_m = B_{m+1}$.
Furthermore, for a state $q_{\it IH}$ with
$I = [A \de \alpha \bul \beta]$ we demand $A=B_1$ if $n>0$.
For a state $q_{\it AH}$ we demand $A=B_1$ if $n>0$.
(Strictly speaking, states $q_{\it AH}$ and $q_{\it IH}$, with 
$|H|<d-1$ 
and 
$I= [A \de \alpha \bul \beta]$,
will only be needed if
$A_{|H|}$ is the start symbol in the case $|H|>0$, or if
$A$ is the start symbol in the case $H=\ep$.)

The transitions of the automaton that pertain to terminals in
right-hand sides of rules
are very similar to those in the case of the unparameterized
method: For a state $q_{\it IH}$ with $I$ of the form
$[A\de \alpha \bul a \beta]$, we create a transition $(q_{\it IH},a,q_{I'H})$, 
with $I'=[A\de \alpha a \bul \beta]$.

Similarly, we create epsilon transitions that connect
left-hand sides
and right-hand sides of rules: For each state $q_{\it AH}$
there is a transition $(q_{\it AH},\ep,q_{\it IH})$ for each item
$I=[A \de\ \bul\alpha]$, for some $\alpha$,
and for each state of the form $q_{\it I'H}$,
with $I'=[A \de \alpha\bul]$, there is a transition
$(q_{\it I'H},\ep,q'_{\it AH})$.

For transitions that pertain to nonterminals in the right-hand sides
of rules, we need to manipulate the histories.
For a state $q_{\it IH}$ with $I$ of the form
$[A\de \alpha \bul B \beta]$, we create two epsilon transitions. One is
$(q_{\it IH},\ep,q_{\it BH'})$, where $H'$ is defined to be ${\it IH}$ if 
$|{\it IH}|<d$, and to be the first $d-1$ items of ${\it IH}$ otherwise.
Informally, we extend the history by the item $I$ representing the rule position
that we have just come
from, but the oldest information in the history
is discarded if the history becomes too long.
The second transition is $(q'_{BH'},\ep, q_{\it I'H})$, 
with $I'=[A\de \alpha B \bul \beta]$.

If the start symbol is $S$, the initial state is $q_S$
and the final state is $q'_S$ (after the symbol $S$ in the subscripts
we find an empty lists of items). Note that
the parameterized method with $d=1$ concurs with
the unparameterized method, since the lists of items then remain
empty.

\begin{figure*}[tb]
\begin{center}
\psfig{figure=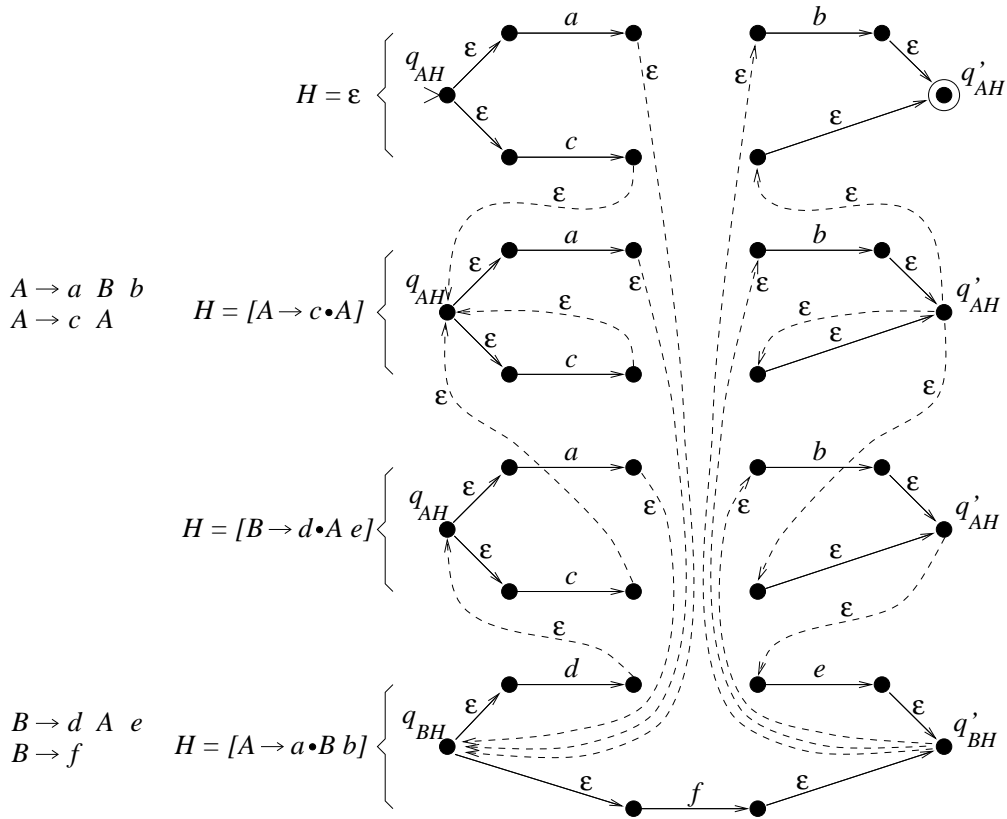}
\end{center}
\caption{Application of the parameterized RTN method with $d=2$.
We again assume $A$ is the start symbol. States $q_{\it IH}$ have not
been labelled in order to avoid cluttering the picture.}
\label{param}
\end{figure*}

An example with parameter $d=2$ is given in
Figure~\ref{param}. 
For the unparameterized method,
each $I=[A \de \alpha\bul\beta]$ corresponded
to one state (Figure~\ref{rtn}). Since reaching $A$ can have
three different histories of length shorter than 2
(the empty history, since $A$ is start
symbol; the history of coming from the rule position given by item
$[A\de c \bul A]$; and, the history of
coming from the rule position given by item
$[B\de d \bul A e]$), in Figure~\ref{param}
we now have three states of the form
$q_{\it IH}$ for each $I=[A \de \alpha\bul\beta]$, as well as three states
of the form $q_{\it AH}$ and $q'_{\it AH}$.

The higher we choose $d$, the more precise the approximation is, since
the histories allow the automaton to simulate part of the 
mechanism of recursion from the original grammar, and 
the maximum length of the histories corresponds
to the number of levels of recursion that can be simulated accurately.

\subsection{Refinement of RTN superset approximation}
\label{approx:gr}

We rephrase the method by \namecite{GR97} as follows.
First, we construct the approximating finite automaton
according to the unparameterized RTN method above.
Then an additional mechanism is 
introduced that ensures for each rule $A\de X_1\cdots X_m$
separately that
the list of visits to the states $q_0,\ldots, q_m$ 
satisfies some reasonable
criteria: a visit to $q_i$, with $0\leq i<m$, should be followed by one to
$q_{i+1}$ or $q_0$. The latter option amounts to a nested incarnation
of the rule. There is a complementary condition for what should
precede a visit to $q_i$, with $0< i\leq m$.

Since only pairs of consecutive visits to states from
the set $\{q_0,\ldots,q_m\}$ are considered,
finite-state techniques
suffice to implement such conditions. This can be realized by attaching
histories to the states as in the case of the parameterized RTN method above,
but now each history is a set rather than a list, and 
can contain at most one item $[A\de\alpha\bul\beta]$
for each rule $A\de\alpha\beta$.
As reported by \namecite{GR97} 
and confirmed by our own experiments, the 
nondeterministic
finite automata resulting from this method 
may be quite large, even for small grammars.
The explanation is that the number of such histories is exponential
in the number of rules.

We have refined the method with respect to the original publication
by applying the construction separately for each
nonterminal in a
set $N_i$ such that $\recursive(N_i)=\self$.

\subsection{Subset approximation by transforming the grammar}
\label{approx:sub}

Putting restrictions on spines is another way to obtain a regular
language. Several methods can be defined. The first method we present
investigates spines in a very detailed way.
It eliminates from the language only those sentences for which a
subderivation is required of the form
$B\dm \alpha B \beta$, for
some $\alpha\neq\epsilon$ and $\beta\neq\epsilon$.
The motivation is that
such sentences do not occur frequently in practice,
since these subderivations make it difficult for
people to comprehend them \cite{RE92b}.
Their exclusion 
will therefore not lead to much loss of coverage of typical
sentences, especially for simple application domains.

We express the method in terms of a grammar transformation in
Figure~\ref{expo.alg}. The effect of this transformation is that a nonterminal $A$ is
tagged with a set of pairs $(B,Q)$, where $B$ is a nonterminal occurring higher in the 
spine; for given $B$, at most one such pair $(B,Q)$ can be contained in the set. 
The set $Q$ may contain the element $l$ to indicate that something to the 
left of the part of the spine from $B$ to $A$ was generated. Similarly, 
$r\in Q$ indicates that something to the
right was generated. If $Q=\{l,r\}$, then we have obtained a derivation
$B\dm \alpha A \beta$, for
some $\alpha\neq\epsilon$ and $\beta\neq\epsilon$, and further occurrences of $B$ below
$A$ should be blocked in order to avoid a derivation with self-embedding. 

\begin{figure*}[tb]
\rule{\textwidth}{.5mm}
We are given a grammar $G=(\myterm,N,P,S)$. The following is to be
performed for each set $N_i\in {\cal N}$ such that
$\recursive(N_i)=\self$.
\begin{enumerate}
\item\ For each $A\in N_i$ and each $F\in 2^{(N_i\times 2^{\{l,r\}})}$, 
add the following nonterminal to $N$.
\begin{itemize}
\item $A^{F}$.
\end{itemize}
\item\ For each $A\in N_i$, add the following rule to $P$.
\begin{itemize}
\item $A \de A^{\emptyset}$.
\end{itemize}
\item\ For each $(A\de \alpha_0 A_1\alpha_1A_2\cdots \alpha_{m-1}A_m\alpha_m)\in P$
such that
$A,A_1,\ldots,A_m\in N_i$
and no symbols from $\alpha_0,\ldots,\alpha_m$
are members of $N_i$, and each $F$ such that $(A,\{l,r\})\notin F$, 
add the following rule to $P$.
\begin{itemize}
\item $A^{F} \de \alpha_0 A_1^{F_1}\alpha_1\cdots A_m^{F_m}\alpha_m$,
where, for $1\leq j \leq m$,
\begin{itemize}
\item $F_j = \{ (B,Q\cup Q_l^j\cup Q_r^j)\ |\ (B,Q)\in F'\}$;
\item $F'= F\cup \{ (A,\emptyset) \}$ if $\neg\exists Q [(A,Q)\in F]$, and $F'= F$
otherwise;
\item $Q_l^j = \emptyset $ if $\alpha_0A_1\alpha_1\cdots A_{j-1}\alpha_{j-1}=\ep$, and
	$Q_l^j = \{l\}$ otherwise;
\item $Q_r^j = \emptyset $ if $\alpha_jA_{j+1}\alpha_{j+1}\cdots A_m\alpha_m=\ep$, and
        $Q_r^j = \{r\}$ otherwise.
\end{itemize}
\end{itemize}
\item\ Remove from $P$ the old rules of the form $A\de\alpha$, where $A\in N_i$.
\item\ Reduce the grammar.
\end{enumerate}
\vspace{-.5cm}
\rule{\textwidth}{.5mm}
\vspace{-.2cm}
\caption{Subset approximation by transforming the grammar.}
\label{expo.alg}
\end{figure*}

\begin{figure*}[tb]
\ \psfig{figure=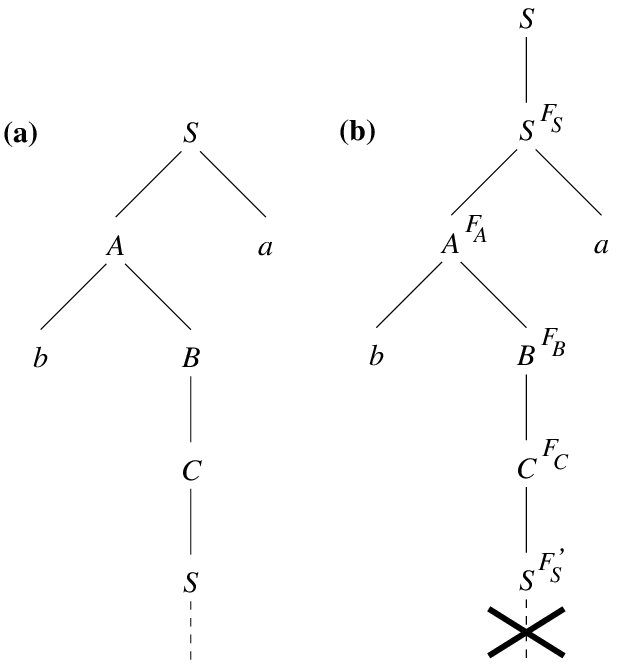}
\hspace{.3cm}
\begin{minipage}[b]{2.6cm}
\begin{eqnarray*}
F_S &=& \emptyset \\[0.6cm]
F_A &=& \{(S,\{r\})\} \\[0.6cm]
F_B &=& \{(S,\{l,r\}),(A,\{l\})\} \\[0.6cm]
F_C &=& \{(S,\{l,r\}),(A,\{l\}),(B,\emptyset)\} \\[0.6cm]
F_S' &=& \{(S,\{l,r\}),(A,\{l\}),(B,\emptyset),(C,\emptyset)\} \\[.5cm]
\end{eqnarray*}
\end{minipage}
\caption{A parse tree in a self-embedding grammar (a), and the
corresponding parse tree in the transformed grammar (b), for
the transformation from Figure~\protect\ref{expo.alg}.
For the moment we ignore step~5 of Figure~\protect\ref{expo.alg}, i.e.\ 
reduction of the transformed grammar.}
\label{seblock}
\end{figure*}

An example is given in Figure~\ref{seblock}. The original grammar is implicit in the
depicted parse tree on the left, and contains at least the rules $S\de A\ a$,
$A \de b\ B$, $B\de C$ and $C\de S$.
This grammar
is self-embedding, since we have a subderivation $S\dm b S a$.
We explain how $F_B$ is obtained from $F_A$ in the rule
$A^{F_A}\de b\ B^{F_B}$. We first construct $F'=\{(S,\{r\}),(A,\emptyset)\}$ 
from $F_A= \{(S,\{r\})\}$ by adding $(A,\emptyset)$, since no
other pair of the form $(A,Q)$ was already present.
To the left of the occurrence of $B$ in the original rule
$A \de b\ B$ we find a non-empty string $b$.
This means that we have to add $l$ to all second components
of pairs in $F'$, which gives us $F_B=\{(S,\{l,r\}),(A,\{l\})\}$.

In the transformed grammar, the lower occurrence of $S$ in the tree
is tagged with the set $\{(S,\{l,r\}),(A,\{l\}),(B,\emptyset),(C,\emptyset)\}$.
The meaning is that higher up in the spine, we will find the nonterminals
$S$, $A$, $B$ and $C$. The pair $(A,\{l\})$ indicates that since we saw $A$
on the spine, something to the left has been generated, viz.\ $b$. 
The pair $(B,\emptyset)$
indicates that nothing either to the left or to the right has been
generated since we saw $B$. The pair $(S,\{l,r\})$ indicates that both
to the left and to the right
something has been generated (viz.\ $b$ on the left and $a$ on the right).
Since this indicates that an
offending subderivation $S\dm \alpha S\beta$
has been found, further completion of
the parse tree is blocked: the transformed grammar
will not have any rules with left-hand side
$S^{\{(S,\{l,r\}),(A,\{l\}),(B,\emptyset),(C,\emptyset)\}}$.
In fact, after the grammar is reduced, any parse tree that is constructed
cannot even contain any longer a
node labelled by $S^{\{(S,\{l,r\}),(A,\{l\}),(B,\emptyset),(C,\emptyset)\}}$,
or {\em any\/} nodes with labels of the form $A^F$ such that $(A,\{l,r\})\in F$.

One could generalize this approximation in such a way that not all self-embedding
is blocked, but only self-embedding occurring, say, twice in a row, in the 
sense of
a subderivation of the form
$A\dm \alpha_1 A \beta_1 \dm \alpha_1\alpha_2 A \beta_2\beta_1$.
We will not do so here, because already for the basic case above,
the transformed grammar can be huge due to the high number of 
nonterminals of the form $A^F$ that may result; the number of such nonterminals
is exponential in the size of $N_i$.

We therefore present, in Figure~\ref{expo3.alg}, 
an alternative approximation that has a lower complexity. By parameter $d$, it restricts
the number of rules along a spine that may generate something to the left and to the right.
We do however not restrict pure left recursion and pure right recursion. Between
two occurrences of an arbitrary rule, we allow left recursion followed by right recursion (which leads
to tag $r$ followed by tag ${\it rl}$), or right recursion followed by left recursion (which leads
to tag $l$ followed by tag ${\it lr}$).

\begin{figure*}[tb]
\rule{\textwidth}{.5mm}
We are given a grammar $G=(\myterm,N,P,S)$. The following is to be
performed for each set $N_i\in {\cal N}$ such that
$\recursive(N_i)=\self$. The value $d$ stands for the maximum number of
unconstrained rules along a spine,
possibly alternated with a series of 
left-recursive rules followed by a series of right-recursive rules, or vice versa.
\begin{enumerate}
\item\ For each $A\in N_i$, each $Q\in \{\top,l,r, {\it lr}, {\it rl},\bot\}$, and each 
$f$ such that $0\leq f \leq d$,
add the following nonterminals to $N$.
\begin{itemize}
\item $A^{Q,f}$.
\end{itemize}
\item\ For each $A\in N_i$, add the following rule to $P$.
\begin{itemize}
\item $A \de A^{\top,0}$.
\end{itemize}
\item\ For each $A\in N_i$ and $f$ such that $0\leq f\leq d$,
add the following rules to $P$.
\begin{itemize}
\item $A^{\top,f} \de A^{l,f}$.
\item $A^{\top,f} \de A^{r,f}$.
\item $A^{l,f} \de A^{{\it lr},f}$.
\item $A^{r,f} \de A^{{\it rl},f}$.
\item $A^{{\it lr},f} \de A^{\bot,f}$.
\item $A^{{\it rl},f} \de A^{\bot,f}$.
\end{itemize}
\item\ For each $(A\de B \alpha)\in P$
such that
$A,B\in N_i$
and no symbols from $\alpha$
are members of $N_i$, each $f$ such that $0\leq f \leq d$,
and each $Q\in\{r,{\it lr}\}$,
add the following rule to $P$.
\begin{itemize}
\item $A^{Q,f} \de B^{Q,f}\alpha$.
\end{itemize}
\item\ For each $(A\de \alpha B)\in P$
such that
$A,B\in N_i$
and no symbols from $\alpha$
are members of $N_i$, each $f$ such that $0\leq f \leq d$,
and each $Q\in\{l,{\it rl}\}$,
add the following rule to $P$.
\begin{itemize}
\item $A^{Q,f} \de \alpha B^{Q,f}$.
\end{itemize}
\item\ For each $(A\de \alpha_0 A_1\alpha_1A_2\cdots \alpha_{m-1}A_m\alpha_m)\in P$
such that
$A,A_1,\ldots,A_m\in N_i$
and no symbols from $\alpha_0,\ldots,\alpha_m$
are members of $N_i$, and each $f$ such that $0\leq f \leq d$,
add the following rule to $P$, provided $m=0 \vee f < d$.
\begin{itemize}
\item $A^{\bot,f} \de \alpha_0 A_1^{\top,f+1}\alpha_1\cdots A_m^{\top,f+1}\alpha_m$.
\end{itemize}
\item\ Remove from $P$ the old rules of the form $A\de\alpha$, where $A\in N_i$.
\item\ Reduce the grammar.
\end{enumerate}
\vspace{-.5cm}
\rule{\textwidth}{.5mm}
\vspace{-.2cm}
\caption{A more simple subset approximation by transforming the grammar.}
\vspace{-0.3cm}
\label{expo3.alg}
\end{figure*}

\begin{figure*}[tb]
\hspace*{\fill}
\psfig{figure=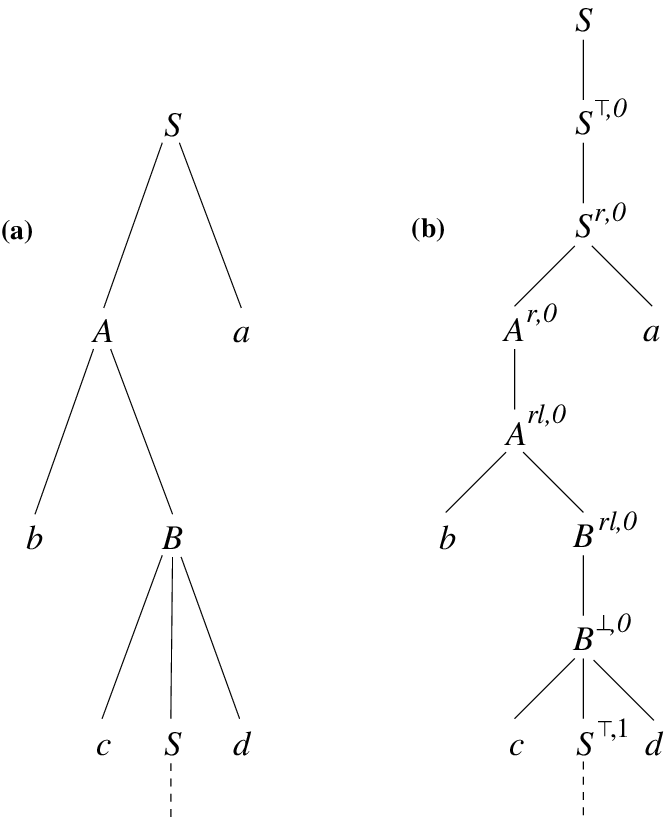}
\hspace*{\fill}
\caption{A parse tree in a self-embedding grammar (a), and the
corresponding parse tree in the transformed grammar (b), for the
simple subset approximation from Figure~\ref{expo3.alg}.}
\label{block}
\end{figure*}

An example is given in Figure~\ref{block}. As before, the rules of the 
grammar are implicit in the depicted parse tree.
At the top of the
derivation we find $S$. In the transformed grammar, we first have to apply
$S \de S^{\top,0}$.
The derivation starts with a
rule $S\de A\ a$, which generates a string (viz.\ $a$) to the right of a
nonterminal (viz.\ $A$). Before we
can apply zero or
more of such rules,
we first have to apply a unit rule $S^{\top,0}\de S^{r,0}$
in the transformed grammar. For zero or more rules that subsequently 
generate something on the left, such as $A \de b\ B$, 
we have to obtain a superscript
containing {\it rl}, and in the example this is done by
applying $A^{r,0}\de A^{rl,0}$. 
Now we are finished with pure left recursion and
pure right recursion, and apply $B^{rl,0}\de B^{\bot,0}$.
This allows us to apply one unconstrained rule, which
appears in the transformed grammar as
$B^{\bot,0}\de c\ S^{\top,1}\ d$. 

Now the counter 
$f$ has been increased from $0$ at the start of the subderivation to $1$ at the end.
Depending on the value $d$ that we choose, we cannot build
derivations by repeating subderivation $S \dm\ b\ c\  S\  d\  a$
an unlimited number of times: at some point the counter will exceed $d$.
If we choose $d=0$, then already the derivation at Figure~\ref{block}~(b) 
is not possible anymore, since no nonterminal in the
transformed grammar would contain 1 in its superscript.

Because of the demonstrated increase of the counter $f$,
this transformation is guaranteed to remove
self-embedding from the grammar. However, it is not as selective
as the transformation we saw before, in the sense that it may also block
subderivations that are not of the form $A\dm \alpha A \beta$.
Consider for example the subderivation from Figure~\ref{block}, but
replacing the lower occurrence of $S$ by any other nonterminal $C$
that is mutually recursive with $S$, $A$ and $B$. Such a 
subderivation $S\dm b\ c\  C\  d\  a$ would also be blocked by
choosing $d=0$. In general, increasing $d$ allows more of such
derivations that are not of the form $A\dm \alpha A \beta$
but also allows more derivations that are of that form. 

The reason for considering this transformation rather than any other
that eliminates self-embedding is purely pragmatic: 
of the many variants we have tried that yield non-trivial
subset approximations, this transformation has the lowest complexity
in terms of the sizes of intermediate structures and of the resulting
finite automata.

In the actual implementation, 
we have integrated 
the grammar transformation and the construction of the finite automaton,
which avoids re-analysis of the grammar to determine 
the partition of mutually recursive
nonterminals after transformation. 
This integration makes use for example of the fact that for fixed $N_i$ and
fixed $f$,
the set of nonterminals of the form $A^{l,f}$, with $A\in N_i$,
is
(potentially) mutually right-recursive. A set of such nonterminals can
therefore be treated as
the corresponding case from Figure~\ref{toFS}, assuming the value {\it right}.

The full formulation of the integrated grammar transformation 
and construction of the
finite automaton is rather long and is therefore not given here.
A very similar formulation, for another grammar transformation,
is given by \namecite{NE98c}.

\subsection{Superset approximation through pushdown automata}
\label{approx:pw}

The distinction between context-free languages and regular
languages can be seen in terms of the distinction between
pushdown automata and finite automata. Pushdown automata 
maintain a stack that is potentially unbounded in height, which 
allows more complex languages to be recognized than in the case of
finite automata. Regular approximation can be achieved by
restricting the height of the stack, as we will see in
Section~\ref{approx:jo}, or by ignoring the distinction between several
stacks when they become too high.

More specifically, the method proposed by \namecite{PE97} first constructs
an LR automaton, which
is a special case of a pushdown automaton. Then, stacks that
may be constructed in the course of recognition of a string are
computed one by one. However, stacks that contain two occurrences of a 
stack symbol are identified with the shorter stack that results by
removing the part of the stack between the two occurrences, including one
of the two occurrences.
This process defines a congruence relation on stacks,
with a finite number of congruence
classes. This congruence relation directly defines a finite automaton:
each class is translated to a unique state of the nondeterministic
finite automaton, shift actions are
translated to transitions labelled with terminals, 
and reduce actions are translated to epsilon transitions.

The method has a high complexity. 
First, construction of an LR automaton, of which the
size is exponential in the size of the grammar,
may be a prohibitively expensive task \cite{NE96}.
This is however
only a fraction of the effort needed to compute the congruence classes,
of which the number 
is in turn exponential in the size of the LR automaton.
If the resulting nondeterministic automaton is determinized,
we obtain a third source of exponential behaviour. 
The time and space complexity of the method are thereby
bounded by a triple exponential function in the size of the grammar.
This theoretical analysis seems to be in keeping with the high costs
of applying this method in practice, as will be shown later in this
article.

As proposed by \namecite{PE97}, our implementation applies the approximation
separately for each nonterminal occurring in a set $N_i$ that reveals
self-embedding.

A different superset approximation based on LR automata was proposed by
\namecite{BA81} and rediscovered by \namecite{HE94}. 
Each individual stack symbol is now translated to one 
state of the nondeterministic finite automaton.
It can be argued theoretically that this approximation differs from the
unparameterized RTN approximation from Section~\ref{approx:ned} only under certain 
conditions that are not likely to occur very often in practice. 
This consideration is
confirmed by our experiments to be discussed later. Our implementation
differs from the original algorithm in that the approximation is 
applied separately for each nonterminal in a set $N_i$ that reveals
self-embedding.

A generalization of this method was suggested by \namecite{BE90}. For a fixed
number $d>0$ we investigate sequences of $d$ top-most elements of 
stacks that may arise in the LR automaton, and we translate these to
states of the finite automaton.
More precisely, we define another congruence relation on stacks, such that we
have one congruence class for each sequence of $d$ stack symbols and this class
contains all stacks that have that sequence as $d$ top-most
elements; we have a separate class for each stack that contains less
than $d$ elements.
As before, each congruence class is translated to one
state of the nondeterministic finite automaton.
Note that the case $d=1$ is equivalent to the
approximation by \namecite{BA81}.

If we replace the LR automaton by a certain type of automaton that
performs top-down recognition, then the method by \namecite{BE90} amounts
to the parameterized RTN method from Section~\ref{approx:ned};
note that the histories from Section~\ref{approx:ned} in fact
function as stacks, the items being the stack symbols.

\subsection{Subset approximation through pushdown automata}
\label{approx:jo}

By restricting the height of the stack of a 
pushdown automaton, one obstructs recognition
of a set of strings in the context-free language, and therefore
a subset approximation results. This idea was proposed by \namecite{KR81},
\namecite{LA87} and
\namecite{PU86}, and was rediscovered by \namecite{BL89} and recently by
\namecite{JO98}. Since the latest publication in this area is more 
explicit in its
presentation, we will base our treatment on
this, instead of going to the historical roots of the method.

One first constructs a
modified left-corner recognizer from the grammar, in the form
of a pushdown automaton. 
The stack height is bounded by a low number; \namecite{JO98} claims
a suitable number would be 5.
The motivation for using the left-corner strategy is that this bound
may not affect the language in case the grammar is not
self-embedding, and thereby the approximation may be exact.
The reason for this is that the height of
the stack maintained
by a left-corner parser is already bounded by a constant in the absence of
self-embedding.

Our own implementation is more refined than the published algorithms mentioned above,
in that it defines a separate
left-corner recognizer for each nonterminal $A$ such that 
$A\in N_i$ and $\recursive(N_i)=\self$, some $i$.
In the construction for one such recognizer, nonterminals that do not
belong to $N_i$ are treated as terminals, as in all other methods discussed here.

\subsection{Superset approximation by $N$-grams}
\label{approx:ngram}

An approximation from \namecite{SE95} can be explained as follows.
Define the set of all terminals reachable from nonterminal $A$ to be
$\myterm_A = \{a\ |\ \exists \alpha,\beta[A\dm \alpha a \beta]\}$.
We now approximate the set of strings derivable from $A$ by 
$\myterm_A^*$, which is the set of strings
consisting of terminals from $\myterm_A$. Our implementation is slightly
more sophisticated by taking $\myterm_A$ to be
$\{X\ |\ \exists B,\alpha,\beta[B\in N_i\wedge 
B\de \alpha X \beta\wedge X\notin N_i]\}$,
for each $A$ such that $A\in N_i$ and $\recursive(N_i)=\self$, for some $i$.
I.e.\ each $X\in \myterm_A$ is a terminal, 
or a nonterminal not in the same set $N_i$ as $A$,
but immediately reachable from set $N_i$, through $B\in N_i$.

This method can be generalized, inspired by \namecite{ST94b},
who derive $N$-gram probabilities from stochastic context-free
grammars. By ignoring the probabilities, each $N=1,2,3,\ldots$
gives rise to a superset approximation
that can be described as follows. The set of strings
derivable from a nonterminal $A$ is approximated
by the set of strings $a_1 \cdots a_n$ such that
\begin{itemize}
\item for each substring $v=a_{i+1} \cdots a_{i+N}$ ($0\leq i \leq n-N$) we have
$A\dm wvy$, for some $w$ and $y$,
\item for each prefix $v=a_1\cdots a_i$ ($0\leq i\leq n$)
such that $i<N$ 
we have $A\dm vy$, for some $y$, and 
\item for each suffix $v=a_{i+1}\cdots a_n$ ($0\leq i\leq n$)
such that $n-i< N$ we have $A\dm wv$, for some $w$.
\end{itemize}
(Again, the algorithms that we actually implemented
are more refined and
take into account the sets $N_i$.)

The approximation from \namecite{SE95} can be seen as the case
$N=1$,
which will henceforth be called the unigram method.
We have also experimented with the cases $N=2$ and $N=3$, 
which will be called the bigram and trigram methods.

\section{Increasing the Precision}
\label{increase}

The methods of approximation described above take as input
the parts of the grammar that pertain to self-embedding. 
It is only for those parts that the language is affected.
This leads us to a way to increase the precision:
before applying any of the above methods of
regular approximation, we first transform the grammar.

This grammar transformation 
copies grammar rules containing recursive nonterminals
and, in the copies, it replaces these nonterminals 
by new non-recursive nonterminals.
The new rules take over part of the roles of the old rules, but
since the new rules do not contain recursion and therefore
do not pertain to self-embedding, they remain unaffected by
the approximation process. 

Consider for example the palindrome grammar from Figure~\ref{pal}.
The RTN method will yield a rather crude approximation, viz.\
the language $\{a,b\}^*$. 
We transform this grammar in order to keep the approximation process
away from the first three levels of recursion. We achieve this
by introducing three new nonterminals $S[1]$, $S[2]$ and $S[3]$, 
and by adding modified copies of the original grammar rules, so that we obtain:
$$\begin{array}{rcl}
S[1] & \de & a\ S[2]\ a\ \ |\ \ b\ S[2]\ b \ \ |\ \ \ep \\
S[2] & \de & a\ S[3]\ a\ \ |\ \ b\ S[3]\ b \ \ |\ \ \ep \\
S[3] & \de & a\ S\ a\ \ |\ \ b\ S\ b \ \ |\ \ \ep \\
S & \de & a\ S\ a\ \ |\ \ b\ S\ b \ \ |\ \ \ep
\end{array}$$
The new start symbol is $S[1]$.

The new grammar generates the same language as before,
but the approximation process
leaves unaffected the nonterminals
$S[1]$, $S[2]$ and $S[3]$ and the rules defining them,
since these nonterminals are not recursive.
These nonterminals amount to the upper three levels of the parse trees,
and therefore the effect of the approximation on the language
is limited to lower levels.
If we apply the RTN method then we obtain 
the language
that consists of (grammatical) palindromes of the form $ww^R$, where
$w\in \{\ep,a,b\} \cup \{a,b\}^2 \cup \{a,b\}^3$,
plus (possibly ungrammatical) strings of the form $w v w^R$, where $w\in\{a,b\}^3$
and $v\in \{a,b\}^*$. ($w^R$ indicates the mirror image of $w$.)

The grammar transformation in its full generality
is given by the following, which is to be applied for
fixed integer $j>0$, which is a parameter of the transformation,
and for each $N_i$ such that $\recursive(N_i)=\self$.

For each nonterminal $A\in N_i$ we introduce
$j$ new nonterminals $A[1],\ldots,A[j]$.
For each $A\de X_1\cdots X_m$ in $P$ such that $A\in N_i$, and $h$
such that $1\leq h \leq j$, we add $A[h]\de X_1'\cdots X_m'$ to $P$, where
for $1\leq k \leq m$:
\begin{eqnarray*}
X_k' & = & X_k[h+1], \mbox{ if } X_k \in N_i \wedge h<j \\
     & = & X_k, \mbox{ otherwise }
\end{eqnarray*}
Further, we replace all rules $A\de X_1\cdots X_m$ such that $A\notin N_i$
by $A\de X_1'\cdots X_m'$, where
for $1\leq k \leq m$:
\begin{eqnarray*}
X_k' & = & X_k[1], \mbox{ if } X_k \in N_i \\
     & = & X_k, \mbox{ otherwise }
\end{eqnarray*}
If the start symbol $S$ was in $N_i$, we let $S[1]$ be the new start symbol.

A second transformation, which shares some characteristics with the one above,
was presented by \namecite{NE97}.
One of the earliest papers suggesting such 
transformations as a way to increase the precision of
approximation is due to \namecite{CU73}, who however only discuss
examples; no general algorithms were defined.

\section{Empirical Results}
\label{compare}

In this section we investigate
empirically how the respective approximation methods
behave on grammars of different sizes and how much the
approximated languages differ from the original context-free languages. This last
question is difficult to answer in a precise way. Both an original context-free
language and an approximating regular language generally consist of an 
infinite number of strings, and the number of strings that are introduced in a superset
approximation or that are excluded in a subset approximation may also be infinite.
This makes it difficult to attach numbers to the ``quality'' of approximations.

We have opted for a pragmatic
approach which does not require investigation of the entire infinite 
languages of the grammar and the finite 
automata, but that looks at a certain finite set of strings that we have taken
from a corpus, as discussed below. For this finite set of strings
we measure the percentage that overlaps with the investigated languages. 

\begin{figure*}[tb]
\hspace{-.5cm}
\psfig{figure=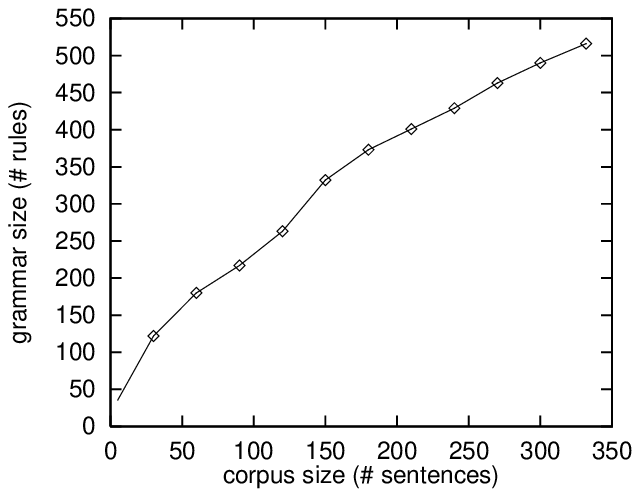}
\hspace{-.5cm}
\psfig{figure=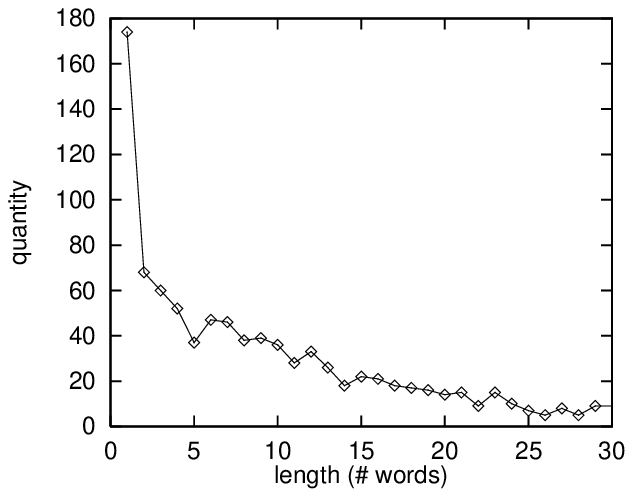}
\caption{The test material. The left-hand curve refers to the
construction of the grammar from 332 sentences, 
the right-hand curve refers to the corpus of 1000 sentences
used as input to the finite automata.}
\label{corgra}
\end{figure*}

For the experiments we have taken context-free grammars for German,
generated automatically from an HPSG and a
spoken-language corpus of 332 sentences. 
This corpus consists of sentences
possessing grammatical phenomena of interest,
manually selected from a larger corpus of actual dialogues.
An HPSG parser was applied
on these sentences, and a form of context-free backbone was selected from
the first derivation that was found. 
(To take the first derivation is
as good as any other strategy, given that we have at present no
mechanisms for relative ranking of derivations.)
The label occurring at a node together
with the sequence of
labels at the daughter nodes was then taken to be a context-free rule.
The collection of such rules for the complete corpus forms a context-free
grammar. Due to the incremental nature of this construction of the grammar,
we can consider the subgrammars obtained after processing the
first $p$ sentences, where $p=1,2,3,\ldots,332$.
See Figure~\ref{corgra} (left) for the relation between $p$ and 
the number of rules of the grammar. The construction is such that 
rules have at most two members in the right-hand side.

As input we consider a set of 1000 sentences, obtained independently
from the 332 sentences mentioned above. These 1000 sentences were
found by having a speech recognizer provide a single hypothesis
for each utterance, where utterances come from actual dialogues.
Figure~\ref{corgra} (right)
shows how many sentences of different lengths
the corpus contains, up to length 30. Above length 25, this number
quickly declines, but still a fair quantity of longer strings
can be found, e.g.\ 11 strings of a length between 51 and 60 words.
In most cases however such long strings are in fact composed 
of a number of shorter sentences.

Each of the 1000 sentences were input in their entirety to the automata,
although in practical spoken-language systems, often
one is not interested in grammaticality of complete utterances,
but one tries to find substrings that
form certain phrases bearing information relevant to the understanding
of the utterance.
We will however not be concerned here with the
exact way such recognition of substrings could be realized
by means of finite automata, 
since this is outside the scope of the present paper.

For the respective methods of approximation we measured
the size of the compact representation of the nondeterministic automaton,
the number of states and the number of
transitions of the minimal deterministic automaton,
and the percentage of sentences that were recognized, in comparison
to the percentage of grammatical sentences. For the compact representation, we
counted the number of lines, 
which is roughly the sum of the numbers of transitions from all
subautomata, not considering about three additional lines 
per subautomaton for overhead.

We have investigated the size of the compact representation because
it is reasonably implementation independent, barring optimizations of the 
approximation algorithms themselves that affect the sizes of the subautomata.
Where we show that for some
method there is a sharp increase in the size of the compact representation for
a small increase in the size of the grammar, this gives us a strong indication 
how difficult it would be to apply
the method to much larger grammars. Note that the size of the compact representation is
a (very) rough indication as to how much effort is involved in 
determinization, minimization, and 
substituting the subautomata into each other.
For determinization and minimization of automata, we have applied programs 
from the FSM library described by \namecite{MO98b}. This library is considered to be
competitive with respect to other tools for processing of finite-state
machines, 
and when the programs cannot determinize or minimize in reasonable time and
space some subautomata constructed by a particular method of approximation,
then this can be regarded to be an indication of the impracticality of the method.

We were not able to compute the compact representation
for all the methods and all the grammars.
Quite problematic proved to be the refined RTN
approximation from
Section~\ref{approx:gr}. 
We were not able to compute the 
compact representation for any of the automatically obtained
grammars in our collection that were self-embedding. 
We therefore eliminated individual rules by hand starting
from the smallest self-embedding grammar in our collection, 
eventually finding grammars 
small enough to be handled by this method. The results are given in 
Table~\ref{ge:numbers}. Note that the size of the compact representation
increases significantly for each additional grammar rule. The sizes of the 
finite automata, after determinization and minimization, remain relatively
small.
\begin{table}[tb]
\tcaption{Size of the compact representation and number of 
states and transitions,
for the refined RTN approximation \protect\cite{GR97}.}
\label{ge:numbers}
\begin{tabular}{rrrr}
grammar size & 
compact repr & 
\# states & 
\# transitions \\
\hline
10 & 133 & 11 & 14 \\
12 & 427 & 17 & 26 \\
13 & 1,139 & 17 & 34 \\
14 & 4,895 & 17 & 36 \\
15 & 16,297 & 17 & 40 \\
16 & 51,493 & 19 & 52 \\
17 & 208,350 & 19 & 52 \\
18 & 409,348 & 21 & 59 \\
19 & 1,326,256 &  21 & 61 \\
\hline
\end{tabular}
\end{table}

Also problematic was the first approximation from Section~\ref{approx:pw},
which was based on LR parsing following \namecite{PE97}.
Already for the grammar of 50 rules, we were not
able to determinize and minimize one of the subautomata according to step~1
of Section~\ref{trees}: we stopped the process after it had become over
600 Megabytes large.
Results as far as we could obtain them are given in Table~\ref{pe:numbers}.
Note the sharp increases in the size of the compact representation,
resulting from small increases, from 44 to 47 and from 47 to 50, 
in the number of rules, and note an accompanying
sharp increase in the size of the finite automaton.
For this method,
we see no possibility to
accomplish the complete approximation process, including
determinization and minimization, for grammars in our collection
that are substantially larger than 50 rules.

\begin{table}[tb]
\tcaption{Size of the compact representation and number of states and transitions,
for the superset approximation based on LR
automata following \protect\namecite{PE97}.}
\label{pe:numbers}
\begin{tabular}{rrrr}
grammar size & compact repr & \# states & \# transitions \\
\hline
35 & 15,921 & 350 & 2,125 \\
44 & 24,651 & 499 & 4,352 \\
47 & 151,226 & 5,112 & 35,754 \\
50 & 646,419&   ? &    ? \\
\hline
\end{tabular}
\end{table}

Since no grammars of interest could be handled by them,
the above two methods will be further left out of consideration.

In the sequel, we refer to the unparameterized and parameterized
approximations based on RTNs
(Section~\ref{approx:ned})
as \mbox{`RTN'} and \mbox{`RTN$d$,'} respectively, for $d=2,3$;
to the subset approximation from Figure~\ref{expo3.alg}
as `Sub$d$,' for $d=1,2,3$; and
to the second and third methods from Section~\ref{approx:pw}, 
which were based on LR parsing
following \namecite{BA81} and \namecite{BE90}, as `LR' and `LR$d$,' 
respectively, for $d=2,3$.
We refer to the 
subset approximation based on left-corner parsing from
Section~\ref{approx:jo} as `LC$d$,'
for the maximal stack height of
$d=2,3,4$; and to the
methods discussed in Section~\ref{approx:ngram}
as `Unigram,' `Bigram' and `Trigram.'

\begin{figure*}[tb]
\psfig{figure=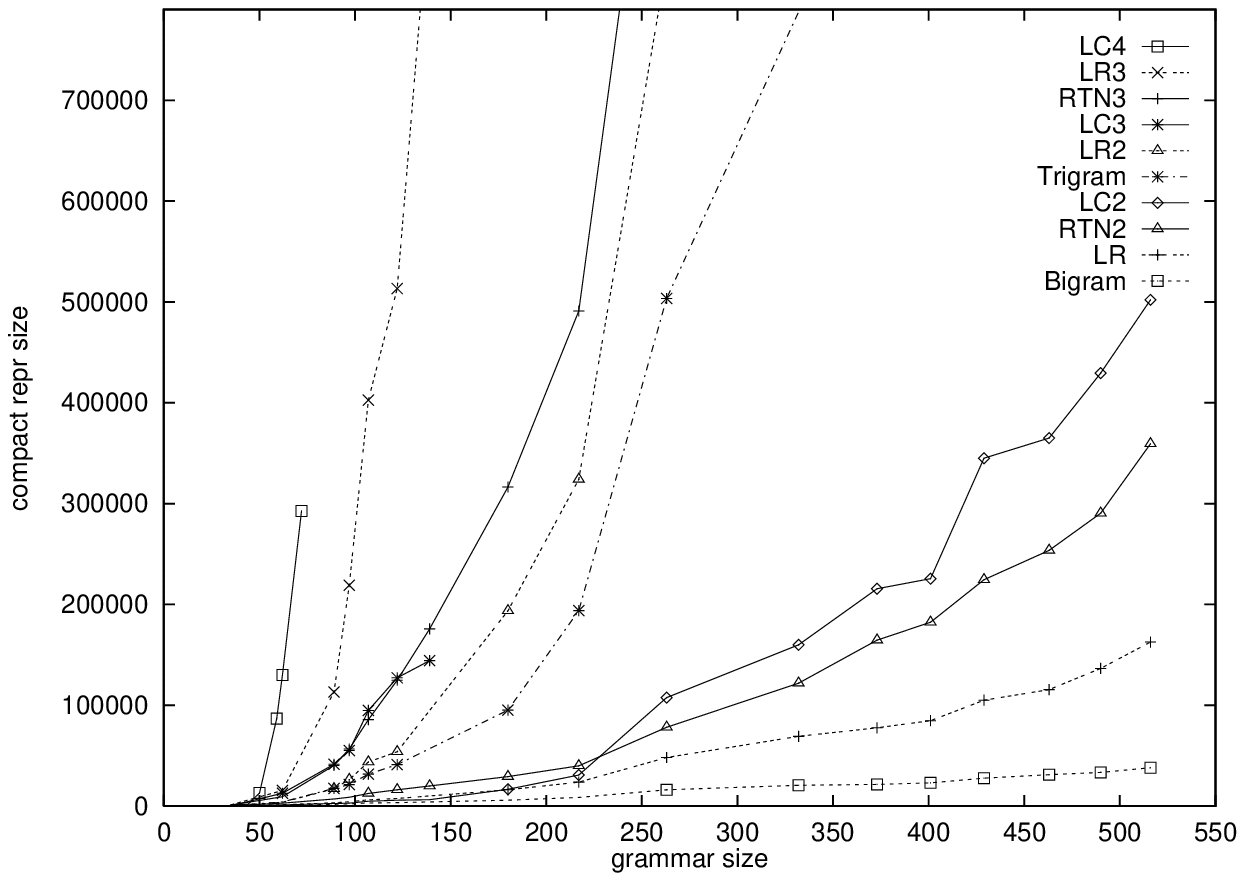} 
\vspace{3ex}
\psfig{figure=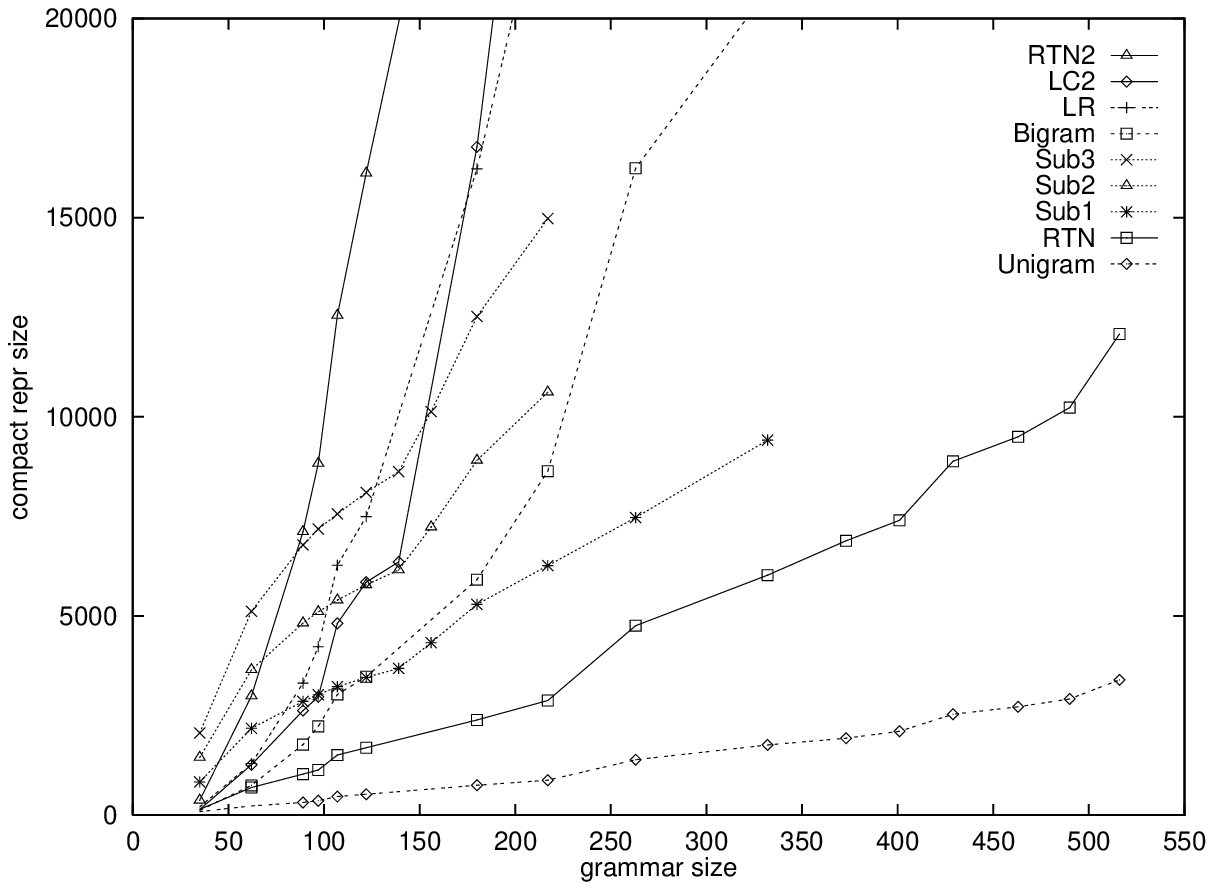}
\caption{Size of the compact representation.}
\label{sizecomp}
\end{figure*}

We first discuss the compact representation of the nondeterministic 
automata.
In Figure~\ref{sizecomp} we use two different scales to be able to represent 
the large variety of values.
For the method Sub$d$, the compact representation is of purely 
theoretical interest for grammars larger than 156 rules in the case of Sub1, 
for those larger than 62 rules in the case of Sub2,
and for those larger than 35 rules in the case of Sub3,
since the minimal deterministic automata could thereafter no longer be
computed with a reasonable bound on resources; we stopped the processes
after they had consumed over 400 Megabytes.
For LC3, LC4, RTN3, LR2 and LR3, 
this was also the case for grammars larger than 139, 62, 156, 217 and 156 rules, 
respectively.
The sizes of the compact representation seem to grow
moderately for LR and Bigram, in the upper panel, yet the sizes are much larger
than those for RTN and Unigram, which are indicated in the lower panel.

\begin{figure*}[tb]
\psfig{figure=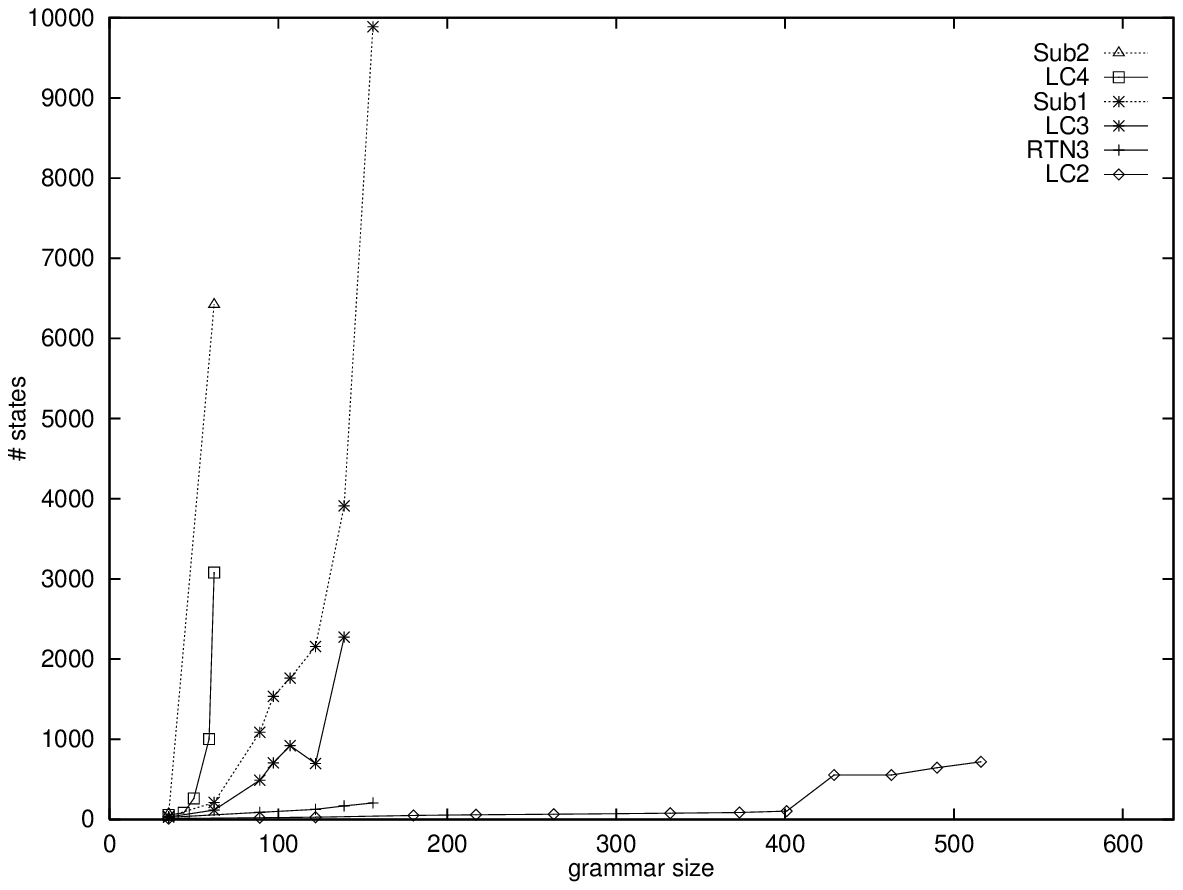}
\vspace{3ex}
\psfig{figure=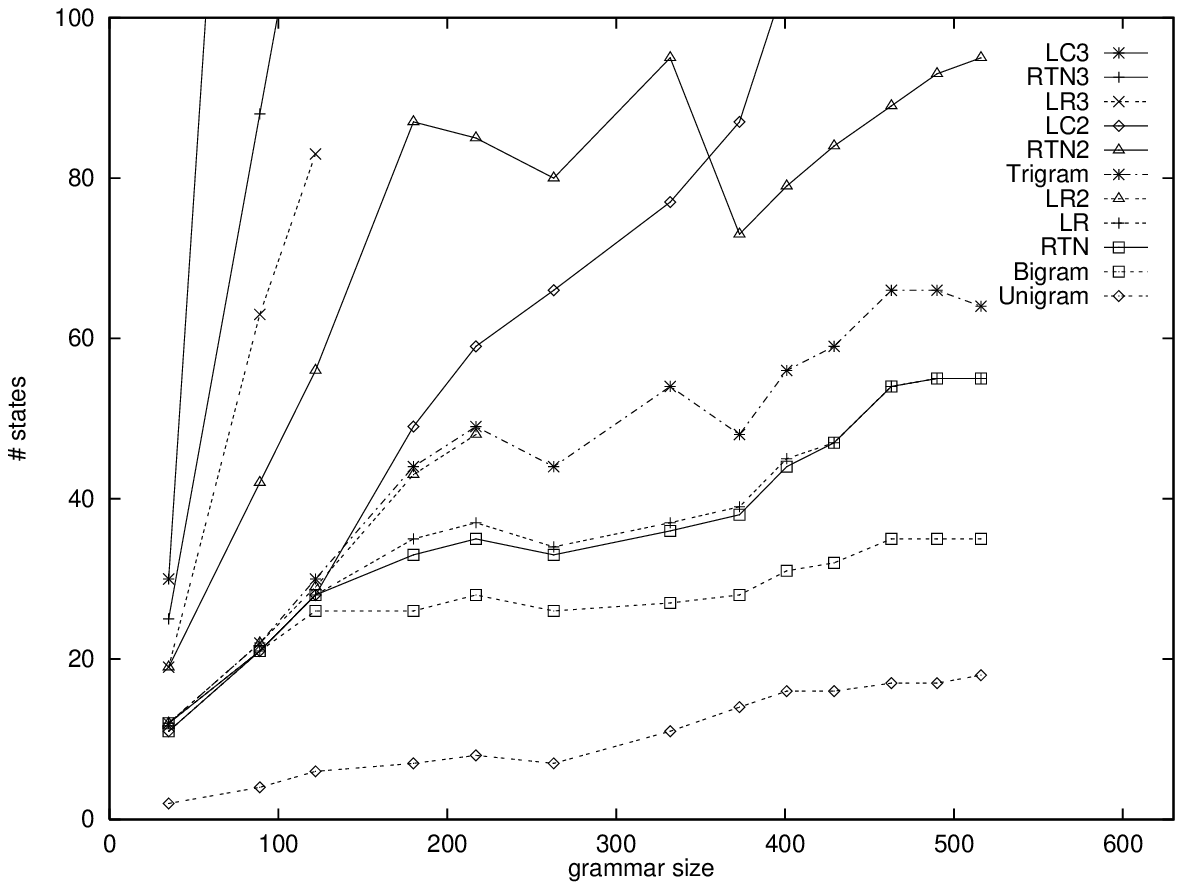}
\caption{Number of states of the determinized and minimized automata.}
\label{nrstates}
\end{figure*}

The numbers of states for the respective methods are given in Figure~\ref{nrstates},
again using two very different scales.
As in the case of the grammars,
the terminals of
our finite automata are parts of speech rather than words.
This means that in general there will be nondeterminism during application 
of an automaton
on an input sentence due to lexical ambiguity.
This nondeterminism can be handled efficiently using tabular techniques 
provided
the number of states is not too high. This consideration favours methods
which produce low numbers of states, such as Trigram,
LR, RTN, Bigram and Unigram.

Note that the numbers of states for LR and RTN differ only very little.
In fact, for some of the smallest and for some of the largest grammars 
in our collection, 
the resulting automata were identical. Remark however that the
intermediate results for LR (Figure~\ref{sizecomp}) are much larger.
It should therefore be concluded that the
``sophistication'' of LR parsing is here merely 
a source of needless inefficiency.

\begin{figure*}[tb]
\psfig{figure=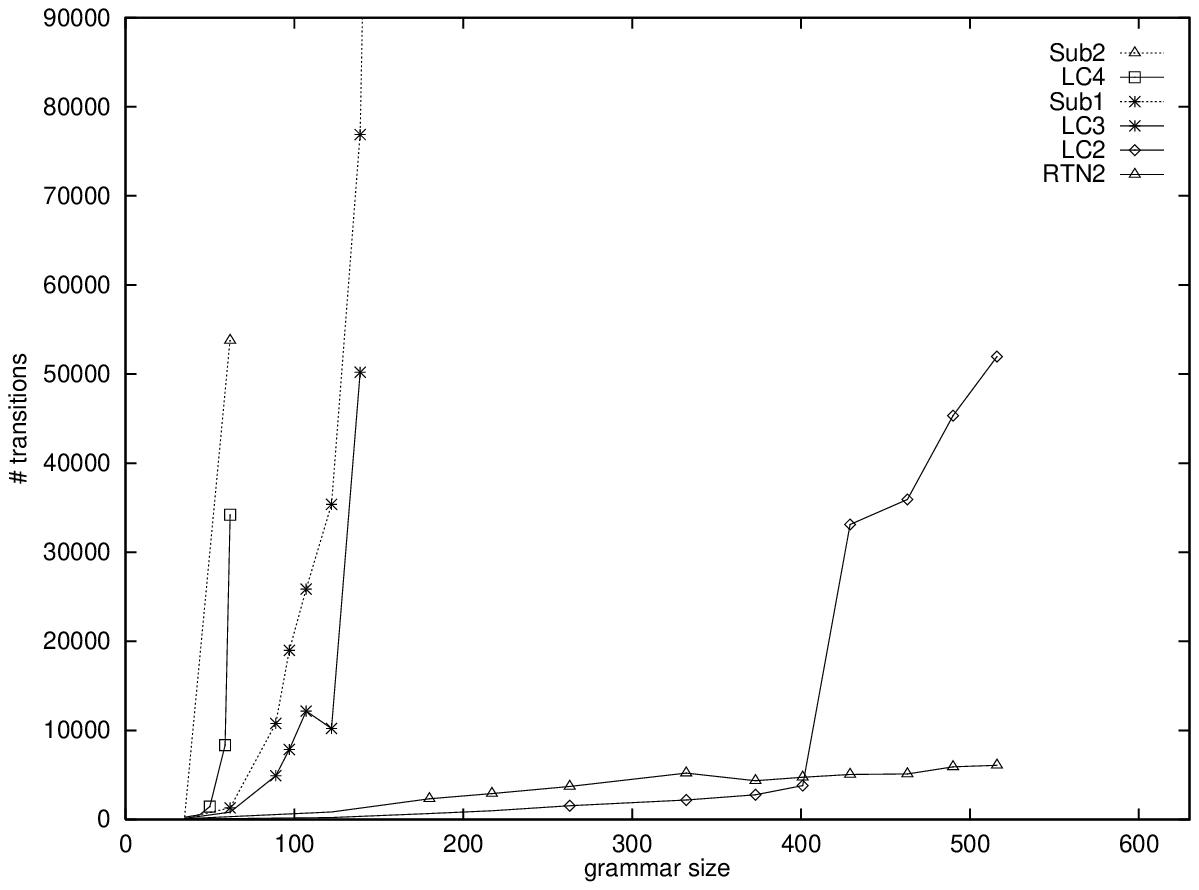}
\vspace{3ex}
\psfig{figure=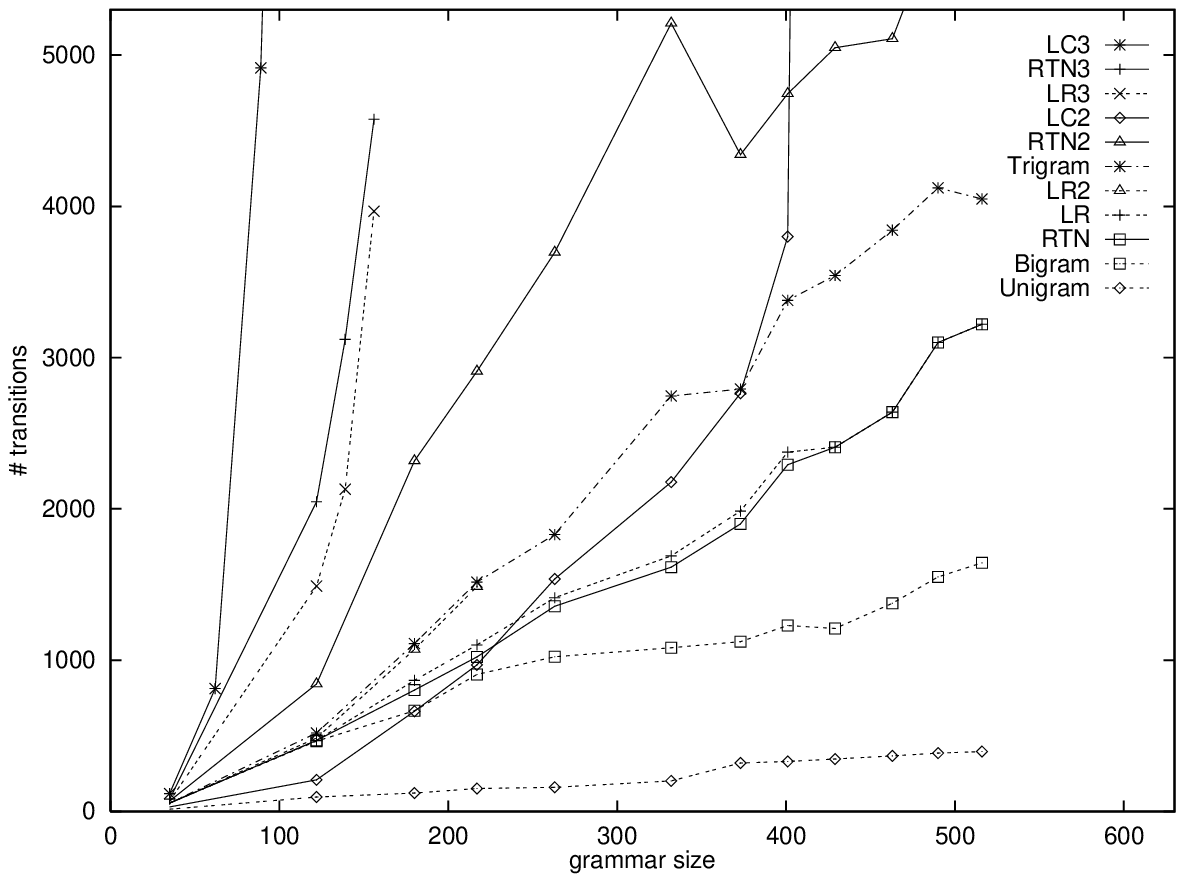}
\caption{Number of transitions of the determinized and minimized automata.}
\label{nrtrans}
\end{figure*}

The numbers of transitions for the respective methods 
are given in Figure~\ref{nrtrans}.
Again note the different scales used in the two panels.
The numbers of transitions roughly correspond to the 
storage requirements for the automata. It can be seen that again
Trigram, LR, RTN, Bigram and Unigram perform well.

\begin{figure*}[tb]
\psfig{figure=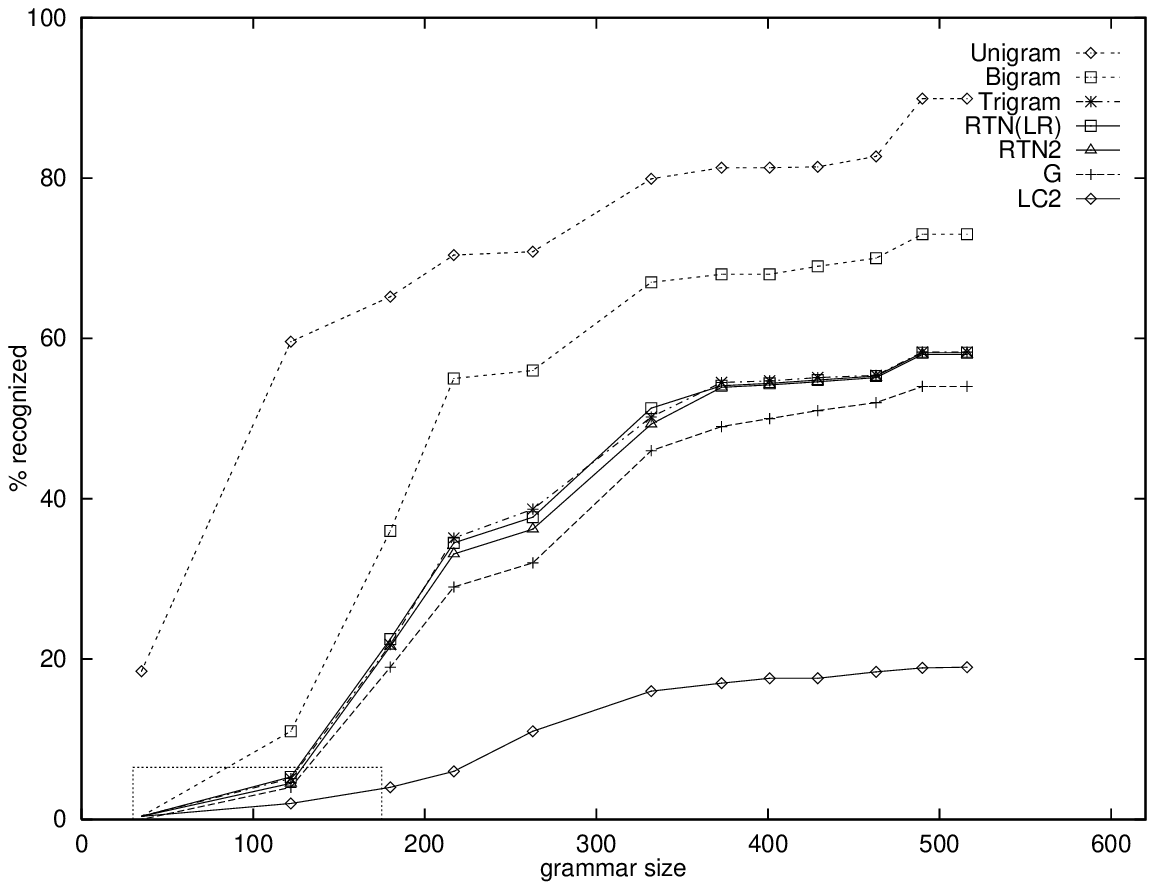}
\vspace{3ex}
\psfig{figure=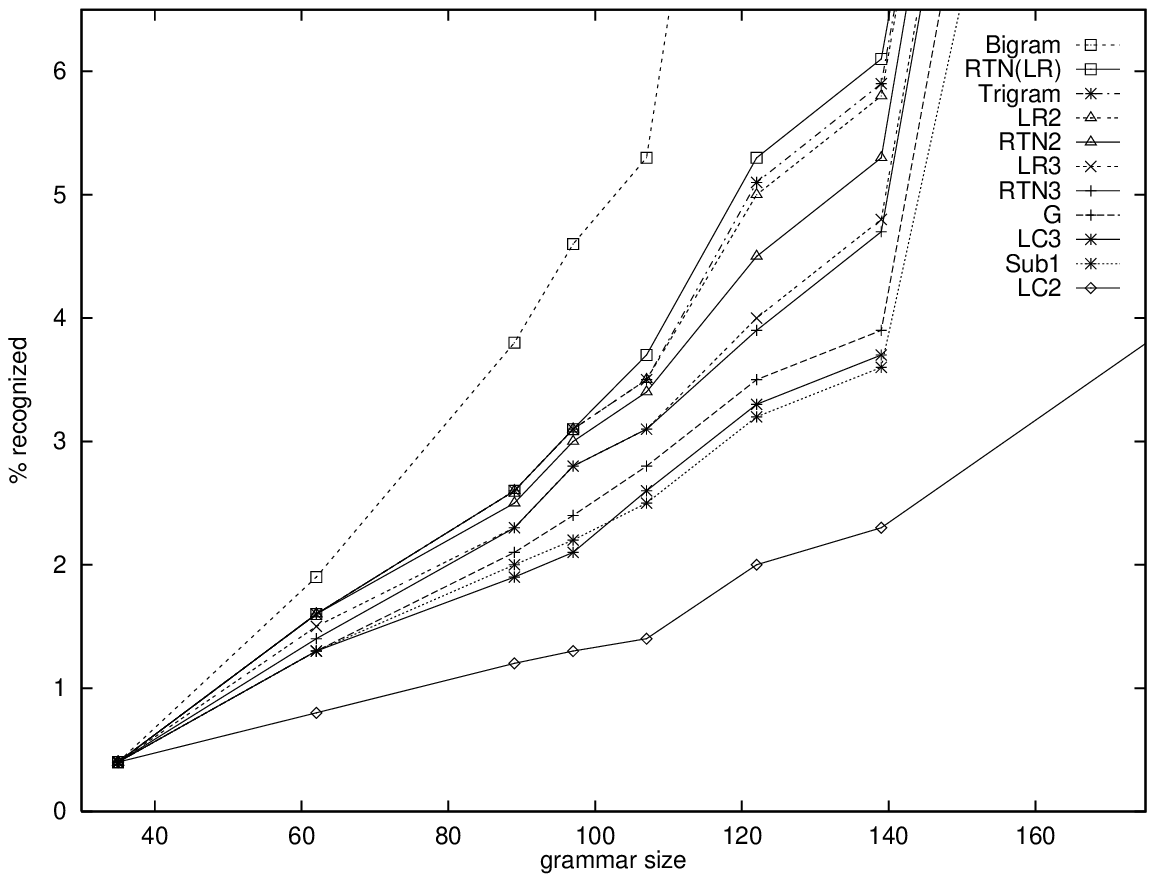}
\caption{Percentage of sentences that are recognized.}
\label{filt}
\end{figure*}

The precision of the respective approximations 
is 
measured in terms of the percentages of sentences in the corpus that are
recognized by the automata, 
in comparison
to the percentage of sentences
that are generated by the grammar,
as presented by
Figure~\ref{filt}.
The lower panel represents an enlargement of a section
from the upper panel. Methods that could only be applied for
the smaller grammars are only presented in the lower panel;
LC4 and Sub2 have been omitted entirely.

The curve labelled G represents the percentage of sentences
that are generated by the grammar.
Note that since all approximation methods compute
either supersets or subsets, 
it cannot occur that
a particular automaton both recognizes
some ungrammatical sentences and rejects some grammatical sentences.

Unigram and Bigram recognize very high percentages of
ungrammatical sentences. Much better results were obtained for RTN.
The curve for LR would not be distinguishable from that for RTN in the 
figure, and is therefore omitted. (For only two of
the investigated grammars was there any difference,
the largest difference occurring for grammar size 217, where
34.1 versus 34.5 percent of sentences were recognized in the cases
of LR and RTN, respectively.)
Trigram remains very close to RTN (and LR);
for some grammars a lower percentage is recognized, for others
a higher percentage is recognized.
LR2 seems to improve slightly over RTN and Trigram, but only for small
grammars is data available,
due to the difficulty of applying the method to larger grammars.
A more substantial improvement is found for 
RTN2. Even smaller percentages are recognized by LR3 and RTN3,
but again, only for small grammars is data available.

The subset approximations
LC3 and Sub1 remain very close to G, but also here 
only data for small grammars is available, since
these two methods could not be applied
on larger grammars.
Although application of LC2 on larger grammars required 
relatively few resources,
the approximation is very crude: only a small percentage of the
grammatical sentences are recognized.

\begin{figure*}[tb]
\hspace{-.5cm}
\psfig{figure=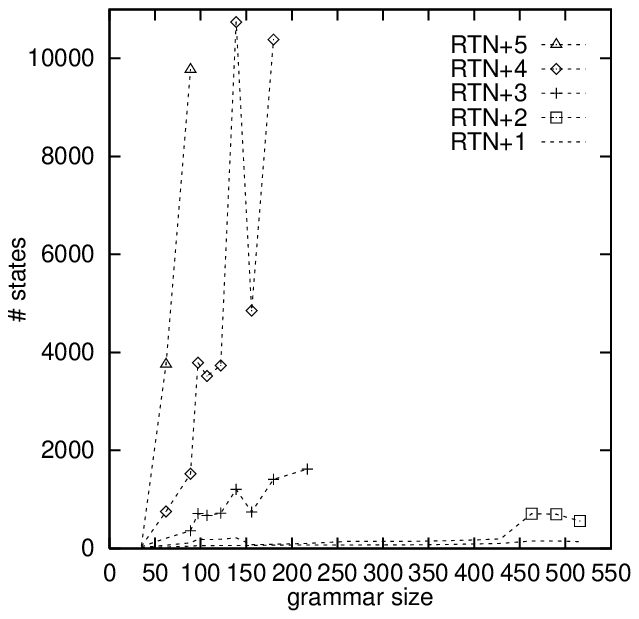}
\hspace{-.5cm}
\psfig{figure=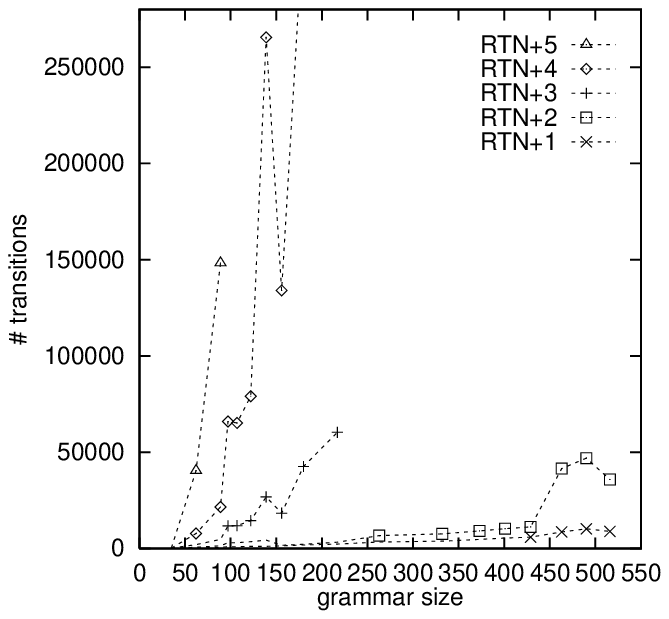}
\caption{Number of states and number of transitions
of the determinized and minimized automata.}
\label{rtnexp}
\end{figure*}

We also performed experiments with the grammar transformation
from Section~\ref{increase}, in combination with the RTN method.
We found that for increasing $j$, the intermediate
automata soon became too large
to be
determinized and minimized, with a bound on the memory
consumption of 400 Megabytes.
The sizes of the automata that we were able to compute
are given in Figure~\ref{rtnexp}. `RTN+$j$,' for $j=1,2,3,4,5$, 
represents the (unparameterized) RTN method in combination 
with the grammar transformation
with parameter $j$. This is not to be confused with the parameterized
`RTN$d$' method.

\begin{figure*}[tb]
\psfig{figure=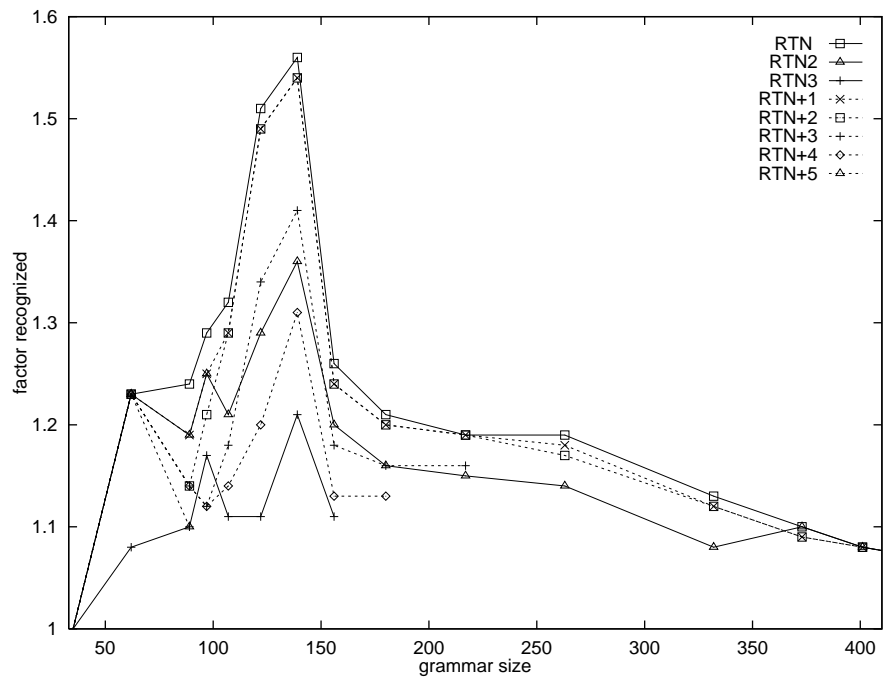}
\caption{Number of recognized sentences divided by number of
grammatical sentences.}
\label{rtnexpacc}
\end{figure*}

Figure~\ref{rtnexpacc} indicates the number of 
sentences in the corpus that are recognized by an automaton
divided by the number of sentences in the corpus 
that are generated by the grammar. 
For comparison, the figure also includes curves for
RTN$d$, where $d=2,3$ (cf.\ Figure~\ref{filt}).
We see that $j=1,2$ has little effect.
For $j=3,4,5$, however, the approximating language
becomes substantially smaller than that in the case of RTN,
but at the expense of large automata.
In particular, if we compare the sizes of the automata for RTN+$j$ in
Figure~\ref{rtnexp} with those for RTN$d$ in
Figures~\ref{nrstates} and~\ref{nrtrans}, then Figure~\ref{rtnexpacc}
suggests the large sizes of the automata for 
RTN+$j$ are not compensated adequately 
by a reduction of the percentage of
sentences that are recognized.
RTN$d$ seems therefore preferable over RTN+$j$.

\section{Conclusions}
\label{conclude}

If we apply the finite automata with the intention of filtering out
incorrect sentences, for example from the output from a speech recognizer,
then it is allowed that
a certain percentage of ungrammatical input is
recognized, since this merely makes 
filtering less effective, but does not affect the functionality of the 
system as a whole, provided we assume that the grammar specifies exactly 
the set of sentences that can be 
successfully handled by a subsequent phase of processing.
Also allowed is that ``pathological'' grammatical
sentences are rejected that seldom occur in practice;
an example are sentences requiring multiple levels of 
self-embedding.

Of the methods we considered
that may lead to rejection of grammatical sentences, i.e.\ the
subset approximations, none seems of much practical value.
The most serious problem is the complexity of the construction of 
automata from the compact representation for large grammars. 
Since the tools we used for obtaining the minimal deterministic automata are
considered to be of high quality, it is 
doubtful that alternative implementations
could succeed on much larger grammars, also considering the sharp increases in the
sizes of the automata
for small increases in the size of the grammar.
Only LC2 could be applied with relatively few resources, but this
is a very crude approximation, which leads to rejection of many 
more sentences than just those requiring self-embedding.

Similarly, some of the superset approximations are not
applicable to large grammars because of the high costs of obtaining 
the minimal deterministic automata.
Some others provide rather large 
languages, and therefore do not allow very effective filtering of ungrammatical
input.
One method however seems to be excellently suited for
large grammars, viz.\ the RTN method;
into consideration come the unparameterized version and the
parameterized version with $d=2$.
In both cases,
the size of the automaton 
grows moderately in the grammar size.
For the unparameterized version, also 
the compact representation grows moderately.
Furthermore, the percentage of recognized sentences remains close to the
percentage of grammatical sentences.
It seems therefore that, under the conditions of our experiments, 
this method is the most suitable regular
approximation that is presently available.

\starttwocolumn

\begin{acknowledgments}
This paper could not have been written without the 
wonderful help of Hans-Ulrich Krieger, who created the
series of grammars that are used in the experiments.
I also owe to him many thanks for countless discussions and for
allowing me to pursue this work.
I am very grateful to the anonymous referees for their inspiring
suggestions.

This work was funded by the German Federal Ministry of
Education, Science, Research and Technology (BMBF) in the
framework of the {\sc Verbmobil} Project under Grant
01 IV 701 V0.
\end{acknowledgments}



\end{document}